%% file: main.tex
\documentclass{bmvc2k}

\input{math_commands.tex}

\usepackage{url}
\usepackage{graphicx}
\usepackage{algorithm}
\usepackage[noend]{algpseudocode}
\usepackage{xcolor}
\usepackage{listings}
\usepackage{wrapfig,lipsum}
\usepackage{multirow}
\usepackage{amssymb} %
\usepackage{pifont} %
\usepackage{diagbox}
\usepackage{tikz}
\usepackage{nicematrix}
\usetikzlibrary{decorations.pathreplacing,calligraphy,calc}
\usepackage{ctable}
\usepackage{tablefootnote}
\usepackage{booktabs}
\usepackage{makecell} 
\usepackage{cleveref}
\crefname{figure}{Fig.}{Figs.}
\Crefname{equation}{Equation}{Equations}

\makeatletter
\renewcommand{\eqref}[1]{\textup{\ref{#1}}}
\makeatother
\usepackage{adjustbox}

\definecolor{codeblue}{rgb}{0,0.4,1}  %
\definecolor{codegray}{rgb}{0,0,0}  %
\definecolor{codepurple}{rgb}{0.58,0,0.82}  %
\definecolor{backcolour}{rgb}{1,1,1}  %

\title{Guiding a diffusion model with itself using sliding windows}

\addauthor{Nikolas Adaloglou}{adaloglo@hhu.de}{1}
\addauthor{Tim Kaiser}{tikai103@hhu.de}{1}
\addauthor{Damir Iagudin}{damir.iagudin@hhu.de}{1}
\addauthor{Markus Kollmann}{markus.kollmann@hhu.de}{1}

\addinstitution{
 Heinrich Heine University \\
 Düsseldorf, Germany
}

\runninghead{Adaloglou et al.}{Guiding a diffusion model using sliding windows}

\begin{document}

\maketitle

\begin{abstract}
Guidance is a widely used technique for diffusion models to enhance sample quality. Technically, guidance is realised by using an auxiliary model that generalises more broadly than the primary model. Using a 2D toy example, we first show that it is highly beneficial when the auxiliary model exhibits similar but stronger generalisation errors than the primary model. Based on this insight, we introduce \emph{masked sliding window guidance (M-SWG)}, a novel, training-free method. M-SWG upweights long-range spatial dependencies by guiding the primary model with itself by selectively restricting its receptive field. M-SWG requires neither access to model weights from previous iterations, additional training, nor class conditioning. M-SWG achieves a superior Inception score (IS) compared to previous state-of-the-art training-free approaches, without introducing sample oversaturation. In conjunction with existing guidance methods, M-SWG reaches state-of-the-art Frechet DINOv2 distance on ImageNet using EDM2-XXL and DiT-XL. The code is available at \url{https://github.com/HHU-MMBS/swg_bmvc2025_official}.
\end{abstract}
\input{text}

\bibliography{ref}

\appendix
\input{appendix}

\end{document}

%% file: math_commands.tex
\usepackage{amsmath,amsfonts,bm}

\def\eqref#1{equation~\ref{#1}}

\def\1{\bm{1}}

\newcommand{\wmg}{WMG}  %
\newcommand{\wg}{weak model guidance}  %
\newcommand{\fdd}{FDD}  %

\newcommand{\e}{\bm \epsilon}  %
\newcommand{\w}{w}  %
\newcommand{\x}{\bm x}  %
\newcommand{\y}{\bm y}
\newcommand{\z}{\bm z}  %

\newcommand{\xmark}{\ding{55}}

\DeclareMathAlphabet{\mathsfit}{\encodingdefault}{\sfdefault}{m}{sl}
\SetMathAlphabet{\mathsfit}{bold}{\encodingdefault}{\sfdefault}{bx}{n}

%% file: text.tex
\section{Introduction}
Diffusion models (DMs) have emerged as a powerful approach for generative tasks, achieving remarkable success in areas such as image synthesis and text-to-image generation \citep{sohldickstein2015deep, ho2020denoising, karras_edm, song2021scorebased, kingma2023variational,saharia2022photorealistic,adaloglou2025rethinking}. Despite their success, DMs often fail to generate high-quality samples \cite{ahsan2024comprehensive} and require guidance techniques to improve visual fidelity (\cref{fig:teaser}). The currently most popular method, classifier-free guidance (CFG), pushes the sample towards regions with higher posterior class probability \cite{tang2025diffusion}. Unlike its predecessor, classifier guidance \cite{dhariwal2021diffusion}, which relies on training an external classifier on labeled noisy images, CFG combines conditional and unconditional denoisers, which can be trained jointly \cite{ho2022classifierfreediffusionguidance}. 

In the following, we denote by $\x$ a noisy image and by $\e(\x, t;c)$ and $\e(\x,t)$ the class conditional and unconditional noise predictors at timestep $t$ of the denoising process \cite{dhariwal2021diffusion}. CFG combines the two noise predictions during sampling using the linear extrapolation scheme 
\begin{equation}
    \tilde \e(\x, t;c) = \e(\x, t;c) + \w[\e(\x, t;c) - \e(\x,t)] \ , \quad \text{with guidance weight } w>0.
\end{equation}
Therefore, CFG can be viewed as a linear extrapolation method \cite{song2021scorebased,bradley2024classifier}. Equivalent extrapolation schemes can be found for all diffusion model formulations, such as target prediction or flow matching \cite{karras_edm,salimans2022progressivedistillationfastsampling}. 

\textbf{Limitations of CFG.} Despite the widespread use of CFG in conditional synthesis \cite{peebles2023scalable_dit}, it comes with notable limitations. First, it increases the training budget. When trained jointly, the unconditional task can consume up to 20\% of the computational cost \cite{ho2022classifierfreediffusionguidance}. Additionally, while CFG reduces class mismatch between samples and conditions of the noise predictor \cite{saharia2022photorealistic}, this benefit comes at the expense of sample diversity, as sampling focuses on regions with high class probability \cite{karras2024guidingdiffusionmodelbad}. Since guidance is a linear extrapolation scheme, the sampling trajectory can overshoot the desired distribution, leading to oversaturated or oversimplified images \cite{kynkäänniemi2024applyingguidancelimitedinterval}. Finally, CFG sampling is restricted to class-conditional generation by design.

\textbf{Weak model guidance.} Recently, a new class of guidance methods has been developed that addresses some of the limitations of CFG. These methods utilize the linear extrapolation scheme of CFG, given by
\vspace{-0.1cm}
\begin{equation} \label{eq:wmg}
    \tilde \e(\x,t) = \e_{\text{pos}}(\x,t) + \w[\e_{\text{pos}}(\x,t) - \e_{\text{neg}}(\x,t)] ,
\end{equation}
in a more generic way. Here, the subscripts \emph{pos} for positive and \emph{neg} for negative refer to the sign of the noise predictors. The idea behind \Cref{eq:wmg} is to extrapolate into high likelihood regions by designing a negative model $\e_{\text{neg}}$ that has an inferior performance compared to a well-performing DM $\e_{\text{pos}}$. Typically, $\e_{\text{neg}}$ is derived from $\e_{\text{pos}}$ by re-training using fewer parameters, architecture-based heuristic manipulations, or shorter training times \cite{ahn2024pag,karras2024guidingdiffusionmodelbad}. We refer to this class of guidance methods as \emph{\wg{} (\wmg)}. While {\wmg} methods seem promising, they often require training additional models, or careful manual selection of specific layers to impair \cite{ahn2024pag}.

\input{fig_wrap/teaser_page_1}

In this paper, we first introduce a toy example to show that (i) WMG samples closer to high likelihood regions than CFG and (ii) guidance works best if $\e_{\text{neg}}$ makes similar errors as $\e_{\text{pos}}$ but stronger. A similar effect can be achieved under increased weight regularization in high-resolution image synthesis. Additionally, we introduce masked sliding window guidance (M-SWG), a novel guidance method designed to upweight long-range dependencies (\cref{fig:teaser}). In contrast to existing guidance techniques \cite{saharia2022photorealistic,ho2022classifierfreediffusionguidance,ahn2024pag,  karras2024guidingdiffusionmodelbad,sadat2024cads,kynkäänniemi2024applyingguidancelimitedinterval,wang2024analysis_cfg}, M-SWG requires neither training nor class conditioning and can be applied to any DM that can process multiple image resolutions. M-SWG achieves a superior IS score without causing oversaturation, while maintaining competitive Frechet distances. Finally, by combining M-SWG with existing guidance techniques \cite{ho2022classifierfreediffusionguidance,karras2024guidingdiffusionmodelbad}, a state-of-the-art Frechet DINOv2 distance (FDD) on ImageNet is achieved using EDM2-XXL and DiT-XL \cite{peebles2023scalable_dit,karras2024analyzingimprovingtrainingdynamics}.

\section{Related work}
Diffusion sampling can be coarsely divided into CFG-based variations and alternative methods. Vanilla CFG can be improved by adding a noise schedule to the condition \cite{sadat2024cads}, introducing monotonically increasing guidance weight schedules \cite{wang2024analysis_cfg}, stepwise intensity thresholding \cite{saharia2022photorealistic}, or applying CFG only at an interval in the intermediate denoising steps \cite{kynkäänniemi2024applyingguidancelimitedinterval}. Despite recent advancements, CFG cannot be used with unlabelled datasets or conditional-only trained denoisers by design. While several sampling methods focus on modifying CFG \cite{sadat2024cads,kynkäänniemi2024applyingguidancelimitedinterval,saharia2022photorealistic,wang2024analysis_cfg, sadat2024eliminating}, training- and condition-free approaches remain an open quest \cite{sadat2024icg}.

Alternative guidance methods can be roughly grouped into three categories: i) \emph{architecture-based impairments}, ii) \emph{image-level manipulations} \cite{hong2023sag}, and iii) inferior capacity models, which we refer to as \emph{weak models}. Architectural impairments typically leverage self-attention maps \cite{ahn2024pag,hong2024smoothed} that are known to capture structure-related information \cite{balaji2022ediff_att,hertz2022prompt_att,nam2024dreammatcher_att,ahn2024pag}. For instance, a handful of manually picked attention maps can be replaced with an identity matrix \cite{ahn2024pag} or filtered using Gaussian smoothing \cite{hong2024smoothed}. Nonetheless, the choice of attention maps depends on the model architecture, which limits the applicability of the method. Finally, architecture-specific impairments can create significant side effects, such as deteriorating the visual structure \cite{hong2024smoothed}. Image-level manipulations such as frequency filtering have been attempted with limited success \cite{hong2023sag}. Hong et al. \cite{hong2023sag} restrict high-frequency filtering to image patches corresponding to high activation areas, achieved by upsampling self-attention maps. However, similar to architecture-based impairments, it requires cherry-picking while being prone to artifacts \cite{hong2024smoothed}. 

Weak models of inferior capacity compared to the positive model can be constructed by limiting the network size (i.e. number of parameters) or its training time \cite{karras2024guidingdiffusionmodelbad}. However, such approaches require additional training from scratch or various weight instances, which are not always available. Concurrent work \cite{alemohammad2024sims} deteriorates the positive model by fine-tuning with its own generated samples to derive the negative model. Different from prior works \cite{karras2024guidingdiffusionmodelbad,alemohammad2024sims}, we aim to systematically study \wmg{} methods, starting with a toy example. Our analysis reveals that the negative model needs to exhibit \emph{a similar modeling error as the positive model, but stronger.} The latter statement has not been explicitly formulated in prior works \cite{ahn2024pag, hong2023sag, karras2024guidingdiffusionmodelbad}.

\section{CFG and \wmg{} are fundamentally different} \label{sec:toy}
\input{figs/toy}

In the following, we introduce a two-dimensional toy example 
(\cref{fig:toymodel}) to visualize the conceptual differences between CFG and \wmg{}. The data distribution of our toy example consists of three data points ${\cal D}=\{\y_1, \y_2, \y_3\}$, where each data point is assigned a different class label $c\in \{1,2,3\}$. 
We can generate trajectories $\x(t)$ by numerically solving the ordinary differential equation (ODE) $d\x(t) = \e^*(\x(t), \sigma(t))d\sigma(t)$, with $\e^*$ the optimal noise predictor in the Bayesian sense, and $\sigma(t)\in [0,\sigma_{max}]$ the current noise level. 

Note that $d\sigma(t)<0$ along a denoising trajectory. It can be shown that for a finite set of $N$ data points, the optimal noise predictor takes the explicit form \cite{karras_edm} 
\begin{equation}
    \e^*(\x, \sigma(t)) = \frac{1}{\sigma(t)} \sum_{i=1}^N (\x-\y_i)p(\y_i|\x,t), \quad p(\y_i|\x,t) = \frac{{\cal N}(\x|\y_i,\sigma(t)^2)}{\sum_{i=1}^N{\cal N}(\x|\y_i,\sigma(t)^2)},
\end{equation}
where ${\cal N}(.|.)$ denotes an isotropic normal distribution. That is, the optimal noise predictor is the normalised residual between $\x$ and the posterior-mean $\sum_{i=1}^N \y_i p(\y_i|\x,t)$ as an estimator for a data point. The optimal noise predictor generates trajectories with endpoints arbitrarily close to one of the data points if the ODE is initialized with $\x_{init}\sim {\cal N}(\x|\bm 0,\sigma_{max}^2)$ and $\sigma_{max}$ is sufficiently large. 

Assume we are given an error-prone noise predictor, $\e_{err}$, whose endpoints are distributed as a superposition of three normal distributions, all with variance $\delta^2$, and each normal distribution centered around a different data point (\cref{fig:toymodel}a). For this special case it can be shown that $\e_{err}(\x,\sigma(t))=\sigma(t)/\tilde\sigma(t) \e^*(\x, \tilde\sigma(t))$, with $\tilde\sigma(t)^2= \sigma(t)^2 + \delta^2$. The question is now, to what extent CFG and \wmg{} can help to shift the trajectory endpoints of the error-prone predictor closer to the data distribution? To answer this question, we introduce the error-prone predictors $\e_{\text{pos}}$, $\e_{\text{neg}}$, and adjust the guidance weight $\w$ of \Cref{eq:wmg} such that the average endpoint error is minimized. 
Using $\e(\bm x, t) := \e(\bm x(t), \sigma(t))$ for brevity, we recover the CFG setting by introducing the class-conditional noise predictor $\e_{\text{pos}}(\x,t) = \e_{err}(\x,t;c)$ and the unconditional noise predictor $\e_{\text{neg}}(\x,t) = \e_{err}(\x,t)$. Likewise we recover the \wmg{} setting by $\e_{\text{pos}}(\x,t) = \e_{err}(\x,t)$ and  $\e_{\text{neg}}(\x,t) = \e_{err}'(\x,t)$, where $\e_{err}'$ differs from $\e_{err}$ by a larger endpoint variance $\delta'>\delta$. The denoising trajectories shown in \cref{fig:toymodel} follow from numerically solving the ODE for $\x(t)$, using \Cref{eq:wmg} instead of the optimal noise predictor. 

\textbf{CFG.} As we use just one datapoint per class, the class conditional noise predictor is given by $\e_{\text{pos}}(\x,t) = \sigma(t)^{-1}(\x-\y_{i_c})$, with $i_c$ the data point index belonging to class $c$, resulting in the special case of straight lines as denoising trajectories (\cref{fig:toymodel}a, CFG). For small guidance weights, the trajectory endpoints match slightly better to the data distribution (\cref{fig:toymodel}b, CFG), but for larger guidance weights, they move away from the data distribution (\cref{fig:toymodel}c, CFG). This behavior is a consequence of the fact that $\e_{\text{pos}}$ drives the trajectory towards the data point with the corresponding class label, whereas $\e_{\text{neg}}$ denoises in the direction of a weighted superposition of all data points. As a result, CFG tends to shift the trajectories away from the center of mass of the data distribution (\cref{fig:toymodel}d, CFG), favoring classifiable endpoints over a fit to the data distribution. This behavior has been described in previous works \cite{ho2022classifierfreediffusionguidance} and scales to higher dimensions, e.g. images, where a better distribution overlap between generated samples and training samples in feature space is only observed for sufficiently small guidance weights.

\textbf{\wmg{}.} In contrast to CFG, increasing the guidance weight for \wmg{} drives trajectory endpoints closer to the data distribution (\cref{fig:toymodel}a-c, \wmg{}), up to the point where trajectories become unstable. The crucial condition for WMG to work well is that, in each step, the prediction of $\e_{\text{pos}}$ is closer to $\e^*$ than the prediction of $\e_{\text{neg}}$, causing the extrapolation to correct the prediction of $\e_{\text{pos}}$ towards the optimal prediction of $\e^*$ (\cref{fig:toymodel}d, \wmg{}). 

\begin{equation}
    \w^*(\x,t) = \frac{\| \e_{\text{pos}}(\x,t) - \e^*(\x,t) \| }{ \| \e_{\text{pos}}(\x,t) - \e_{\text{neg}}(\x,t) \|}
    \label{eq:optimal_weight}
\end{equation}

\input{fig_wrap/swg_diagram}
For the special case that $\e_{\text{pos}}$ and $\e_{\text{neg}}$ exhibit the same error up to a multiplicative factor, it is possible for \wmg{} to closely recover the prediction of the optimal denoiser $\e^*$, under the optimal guidance weight $\w^*(\x,t)$ (\Cref{eq:optimal_weight} and supplementary material). The optimal weight is obviously of little practical use as it requires knowledge about the optimal noise predictor $\e^*$. However, it illustrates that the practical choice of a constant guidance weight is, in general, an \textit{ad hoc} simplification. Surprisingly, we observe excellent, albeit imperfect, matching between endpoints and datapoints for constant but sufficiently large weights (\cref{fig:toymodel}).

\section{Sliding window guidance (SWG)} \label{sec:swg}

Assuming that a significant source of the predictor's error can be attributed to long-range dependencies, we introduce a new guidance method that aims to correct such errors. The idea is to generate $\e_{\text{neg}}$ by restricting the spatial input size ($H \times W$) of $\e_{\text{pos}}$ to a smaller size $k \times l$, with $k<H$ and $l<W$, which induces a defined cut-off for long-range dependencies. We use sliding windows to generate $N$ crops of size $k \times l$, using a fixed stride $s$ per dimension. These crops are independently processed (without rescaling) by $\e_{\text{pos}}$ as illustrated in \cref{fig:swg_diag}. The $N$ predictions of $\e_{\text{pos}}$ are superimposed in the same order and at the same positions, which results in an $H\times W$ output for $\e_{\text{neg}}$. Overlapping pixels are averaged. A pseudo-code for SWG is provided in the supplementary material. We emphasize that using sliding windows is a common technique in high-resolution 2D/3D image processing \cite{cardoso2022monai,jiang2023mistral,shao2024learning,ruhe2024rolling_sliding,bar2023multidiffusion}.

\textbf{Hyperparameters and limitations.} One limitation of SWG is that $\e_{\text{pos}}$ must process inputs of varying image resolution. Nonetheless, current architectures, such as Unets \cite{ho2020denoising} or (diffusion) image transformers \cite{peebles2023scalable_dit}, generally satisfy this requirement. SWG can be implemented in a few lines of code and requires no training-related modifications or class conditioning. Unless otherwise specified, we assume $H=W$ and use crop size $k=l \approx 5/8 \times H$ and $N=4$, which translates into $k=40, s=24$ at $64 \times 64$ resolution. We denote the overlap ratio per dimension as $r=1-\frac{s}{k}=0.4$. 

{\textbf{SWG in latent DMs.} We keep the same design choices for latent DMs \cite{rombach2022high} that operate in $\frac{H}{8} \times \frac{H}{8}$ feature dimensions, where $8$ is the upscaling/downscaling factor. More specifically, SWG is performed in each denoising step in the low-dimensional latent space (i.e. $64\times 64$), and the final output is passed to the decoder. The decoder handles the upscaling to the image dimensions, i.e., from $64 \times 64$ to $512 \times 512$. The rationale behind SWG operating in lower feature dimensions is that latent representations, i.e., of a VAE, preserve to a significant extent the spatial structure of the image \cite{kouzelis2025eq,dieleman2025latents}. To make SWG compatible with DiT, we add positional encoding interpolation to process the crops. The choice of $k\approx \frac{5}{8} H$ is determined by practical considerations of Unets. Specifically, $k$ must satisfy $\frac{k}{2^n} \in \mathbb{N}$, where $n$ is the number of downsampling steps of the Unet, and the diffusion model $\e_{\text{pos}}$ must be able to process multiple image resolutions.}

\textbf{Computational overhead.} The computational overhead of SWG compared to CFG is determined by the overlap ratio $r$ and the architecture. For EDM-S and $r=0.4$, we measured an overhead of less than 30\% compared to CFG based on a naive implementation of SWG. More details are provided in the supplementary material.

\input{merge/merge_1}

\textbf{Masked SWG (M-SWG).} We experimentally observe that applying guidance only to overlapping regions of the sliding windows shows better performance than applying guidance to the whole image. We therefore introduce a binary mask $\mathbf{M}\in \{0,1\}^{H\times W}$ to filter for overlapping regions when applying guidance, $\tilde \e(\x,t) = \e_{\text{pos}}(\x,t) + \w \mathbf{M} \odot [\e_{\text{pos}}(\x,t) - \e_{\text{neg}}(\x,t)]$.

\textbf{Combining SWG with existing methods.} Since SWG and M-SWG are conceptually different from existing techniques, we investigate whether they can enhance existing techniques as a linear combination of their stepwise outcomes \cite{liu2022compositional}. Mathematically,  
\begin{equation}
    \tilde \e(\x,t) = \e_{\text{pos}}(\x,t) + \sum_{i} \w_i[\e_{\text{pos}}(\x,t) - \e^{(i)}_{\text{neg}}(\x,t)],
    \label{eq:combine_guidance}
\end{equation}
where $i$ indicates different guidance methods. Quantitative improvements would suggest a complementary relationship between methods. For instance, combining CFG with SWG would increase noise-condition alignment whilst conforming to long-range dependencies.

\section{Experimental evaluation and discussion}
\textbf{Diffusion guidance baselines.} 
We divide the evaluation into training-free methods (\Cref{tab:swg_results}) and state-of-the-art comparisons (\Cref{tab:combine_swg_cfg_rct_iccv}), where M-SWG is used in conjunction with existing methods (\Cref{eq:combine_guidance}). Regarding training-free methods, we consider reduced training (RT) as a state-of-the-art baseline \cite{karras2024guidingdiffusionmodelbad}, along with sampling without guidance ($w=0$). Nonetheless, publicly available DMs do \emph{not} always offer earlier weight instances such as DiT-XL \cite{peebles2023scalable_dit} and StableDiffusion \cite{rombach2022high}, and thus RT cannot be applied. When combining M-SWG with other non-training-free state-of-the-art methods (\Cref{eq:combine_guidance}), we adopt CFG and the recent approach of reducing training time and learning capacity (\emph{RCT}), also known as \emph{``autoguidance''} \cite{karras2024guidingdiffusionmodelbad}. We emphasize that CFG requires jointly or independently training a conditional and an unconditional denoiser, and it is thus not training-free.

\input{fig_wrap/pareto_study}

\textbf{Implementation details.} The experiments were conducted on 4 Nvidia A100 GPUs with 40GB of VRAM. Following recent works \cite{stein2023exposingflawsgenerativemodel, karras2024analyzingimprovingtrainingdynamics, jayasumana2024rethinking_fid, kynkäänniemi2024applyingguidancelimitedinterval, karras2024guidingdiffusionmodelbad, chong2020effectivelyunbiasedfidinception}, we produce 50K samples for the evaluations and measure the Frechet distance \cite{fid_orig} using features from InceptionV3 and DINOv2 ViT-L \cite{oquab2023dinov2}, denoted by FID and \fdd{}, respectively. We use \fdd{} as our primary metric since it has shown better alignment with human judgment \cite{parmar2022aliasedresizingsurprisingsubtleties,morozov2021self,kynkäänniemi2023role,stein2023exposingflawsgenerativemodel}. We also report the inception score (IS) as a measure of visual quality. We apply SWG and M-SWG to state-of-the-art models, such as DiT-XL \cite{peebles2023scalable_dit}, and EDMv2 \cite{karras2024analyzingimprovingtrainingdynamics}.

Regarding EDM2, we set the EMA-like hyperparameter to $0.1$ to make our analysis more generally applicable. Following Karras et al. \cite{karras2024guidingdiffusionmodelbad}, we use $T/16$ for EDM2-S and $T/3.5$ for EDM2-XXL for the negative model when applying RT and RCT, where $T$ is the training time of the positive model. When applying CFG to EDM2 models, we use a smaller capacity unconditional model (EDM2-XS), which we denote as CFG$^\dagger$. Additional implementation details are provided in the supplementary material.

\textbf{M-SWG versus training-free methods.} In \Cref{tab:swg_results}, we observe significant gains compared to the unguided generation using SWG and M-SWG. M-SWG provides a competitive alternative when previous weight instances, such as with DiT-XL, are unavailable. We believe that the inferior Frechet distances from SWG/M-SWG are attributed to background simplification, which strongly affects the statistics of the image distribution. This is a shared limitation with CFG \cite{ho2022classifierfreediffusionguidance, chung2024cfg++} and can be further investigated in future work. Crucially, by combining RT with M-SWG, we achieve a new state-of-the-art \fdd{} for training-free image synthesis using EDM2-XXL (32.9), which also outperforms CFG (36.7). The gains in \fdd{} do not translate to FID. We hypothesize that the less-trained negative model (RT) makes more errors related to local image details, complementing SWG, and it is best captured in the DINOv2 feature space. Finally, no training-free approach can significantly outperform the unguided EDM2-XXL FID score. This is linked to existing concerns that FID does not reflect gradual improvements \cite{parmar2022aliasedresizingsurprisingsubtleties,kynkäänniemi2023role,stein2023exposingflawsgenerativemodel,jayasumana2024rethinking_fid,adaloglou2025rethinking}, as well as that the EDM2-XXL makes fewer errors that can be corrected with guidance techniques. 

Similar to EDM2, applying M-SWG to DiT is highly beneficial in terms of FID/FDD. To the best of our knowledge, early training weight instances are not publicly available for DiT to benchmark against RT or other WMG variants. The poor performance of the unguided DiT-XL is mainly attributed to the joint training on both conditional and unconditional generation, which may reduce network capabilities and lead to skewed sampling distributions \cite{karras2024guidingdiffusionmodelbad}.

\input{merge/merge_2_tables}

\textbf{State-of-the-art comparison.} \Cref{tab:combine_swg_cfg_rct_iccv} shows consistent improvements when combining M-SWG with other state-of-the-art methods. A large relative \fdd{} gain of 9.6\% is found when combining M-SWG with CFG using DiT-XL. Compared to using M-SWG in conjunction with RT versus RCT, the relative gain decreases from 4.9\% to 1.6\%. This suggests that the model of reduced learning capacity partially accounts for the correction in long-range dependencies. Finally, M-SWG shows significant improvements on the unconditional benchmarks. Future work can explore SWG and M-SWG for text-conditioned DMs.

\textbf{Quality-diversity tradeoffs.} To investigate the achieved tradeoff of M-SWG, we plot $w$ against \fdd{}, IS score, and saturation in \cref{fig:pareto}. Saturation is quantified as the mean channel intensity in HSV color space (S channel), as in \cite{sadat2024eliminating}. While RT and RCT are slightly superior in terms of \fdd{}, M-SWG achieves a better image quality (IS) across guidance scales. Intriguingly, we find that M-SWG does not suffer from oversaturation, unlike the CFG and RCT. A low saturation is preferred because it is closer to the statistics of real images (0.32 on the ImageNet test set). Moreover, the root mean square contrast of the samples (the standard deviation of intensities after greyscale conversion \cite{peli1990contrast,sadat2024eliminating}) demonstrates an even more pronounced disparity between M-SWG and other methods (see supplementary material). {Overall, the smaller FID improvements of M-SWG (e.g. 2.55 versus 2.91 for EDM2-S in \Cref{tab:swg_results}), while achieving a higher IS score than RC, RCT suggests that the merits of M-SWG are more in improving semantic coherence than diversity of images.}

\textbf{Additional experiments.} \Cref{tab:ablation-kernel} shows that several combinations of kernel size and $N$ crops can attain similar results on both FID and \fdd. Intriguingly, while FID prefers higher overlap ratios $r$ (for $N=4)$, \fdd{} shows that the overlap between crops is not required to obtain improvements over the unguided model. This suggests that (i) even without overlapping crops, SWG introduces no severe artifacts w.r.t. generative metrics, and (ii) minimal to no tuning is required as opposed to heuristic model-specific impairments like SAG or PAG \cite{hong2023sag,ahn2024pag}. Moreover, in \Cref{tab:ablation-interval}, we
demonstrate that applying guidance only at an intermediate interval further enhances SWG and M-SWG.

\input{figs/oranges}

\textbf{Same error but stronger in high dimensions.} Motivated by the toy model, we verify its implication in higher dimensions by constructing $\e_{\text{neg}}$ with the same network architecture as $\e_{\text{pos}}$, but with significantly increased weight regularization. This can be implemented by either fine-tuning $\e_{\text{pos}}$, denoted as \emph{Weight decay (WD) fine-tuning}, or by re-training it with very little compute time, which we call \emph{WD with reduced training}. WD increases the model's bias in the context of the bias-variance tradeoff \cite{von2011statistical}. Naive training-free approaches such as dropout or quantization to construct $\e_{\text{neg}}$ have been unsuccessful \cite{karras2024guidingdiffusionmodelbad}. Experimental results are presented in \Cref{tab:verification_edm_wmg_in1k}. Interestingly, WD with reduced training performs comparably with the current state-of-the-art RCT, demonstrating that it is another principled way of constructing $\e_{\text{neg}}$.

\textbf{Qualitative results.} To visualize the differences between the guidance methods, we examine ImageNet-512 samples generated from the same initial noise in \cref{fig:oranges}. All three methods, RCT, SWG, and CFG$^\dagger$, tend to exhibit improved perceptual quality for moderate guidance weights but show reduced variation (SWG, CFG$^\dagger$) or artifacts (RCT) for higher guidance weights. To understand these effects, it is helpful to examine the unguided samples produced by $\e_{\text{neg}}$, which corresponds to setting $w=-1$ in \Cref{eq:wmg}. For example, comparing the first two columns of \cref{fig:oranges}, we see that the image composition (e.g.\ box of fruits) for $\e_{\text{pos}}$ ($w=0$) is only maintained by $\e_{\text{neg}}$ ($w=-1$) for RCT. Consequently, with increasing $w$, RCT meaningfully conserves the image composition, SWG does so slightly, and CFG$^\dagger$ not at all.

\section{Conclusion}
In this work, we identified the general concept of ``same error but stronger'' for designing diffusion guidance methods. We validated this concept using a toy example and experimentally in high dimensions using a negative model with increased weight regularization. Additionally, we presented a novel training-free and condition-free guidance method (M-SWG) that enhances global coherence and consistency of objects. M-SWG can be used as a standalone method or combined with state-of-the-art sampling methods, such as CFG, to yield visually appealing and state-of-the-art results in image synthesis.

\vspace{-0.1cm}
\section*{Acknowledgements}
\vspace{-0.1cm}

Computing resources were provided by the German AI Service Center WestAI under the project name “westai9566”. In particular, the authors would like to express their gratitude to Prof. Dr. Timo Dickscheid and Fritz Niesel for their guidance regarding the computational resources used for the project. The authors also wish to thank Kaspar Senft, Georgios Zoumpourlis, and Felix Michels for their constructive feedback, which greatly enhanced the quality of this manuscript.

%% file: fig_wrap/teaser_page_1.tex
\begin{wraptable}{r}{.47\textwidth}
    \centering \includegraphics[width=1\linewidth]{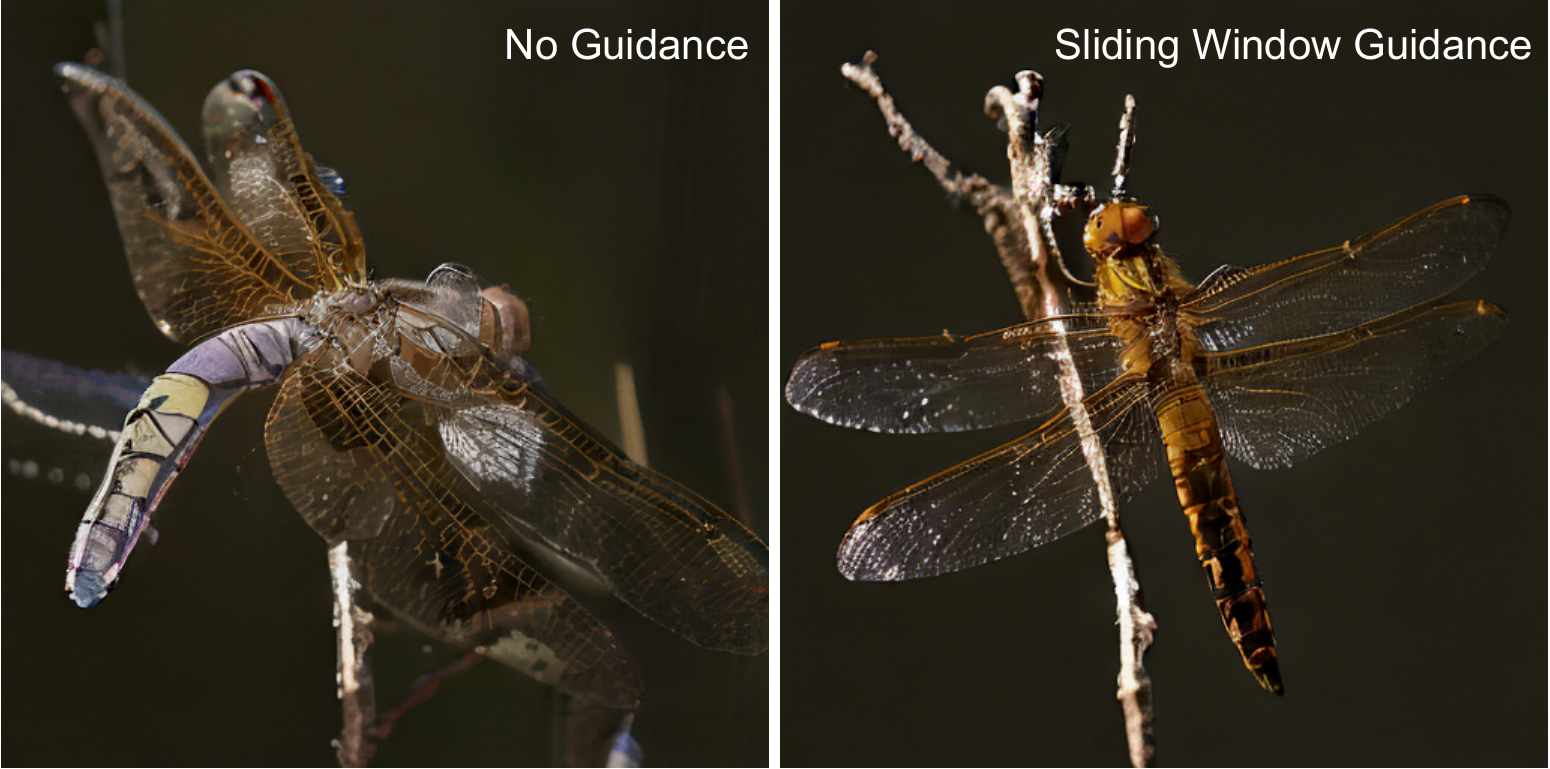} 
    \captionof{figure}{\textbf{Method overview. Left}: Even state-of-the-art diffusion models can fail to generate globally coherent images without guidance. \textbf{Right}: Sliding window guidance (SWG) upweights long-range dependencies and thereby improves global coherence on average.} 
    \label{fig:teaser}
\end{wraptable}

%% file: figs/toy.tex
\begin{figure}
     \centering
     \begin{tikzpicture}
        \node[inner sep=0pt] (grid) at (0,0) {
            \begin{subfigure}[b]{0.26\textwidth}
                 \includegraphics[scale=0.36]{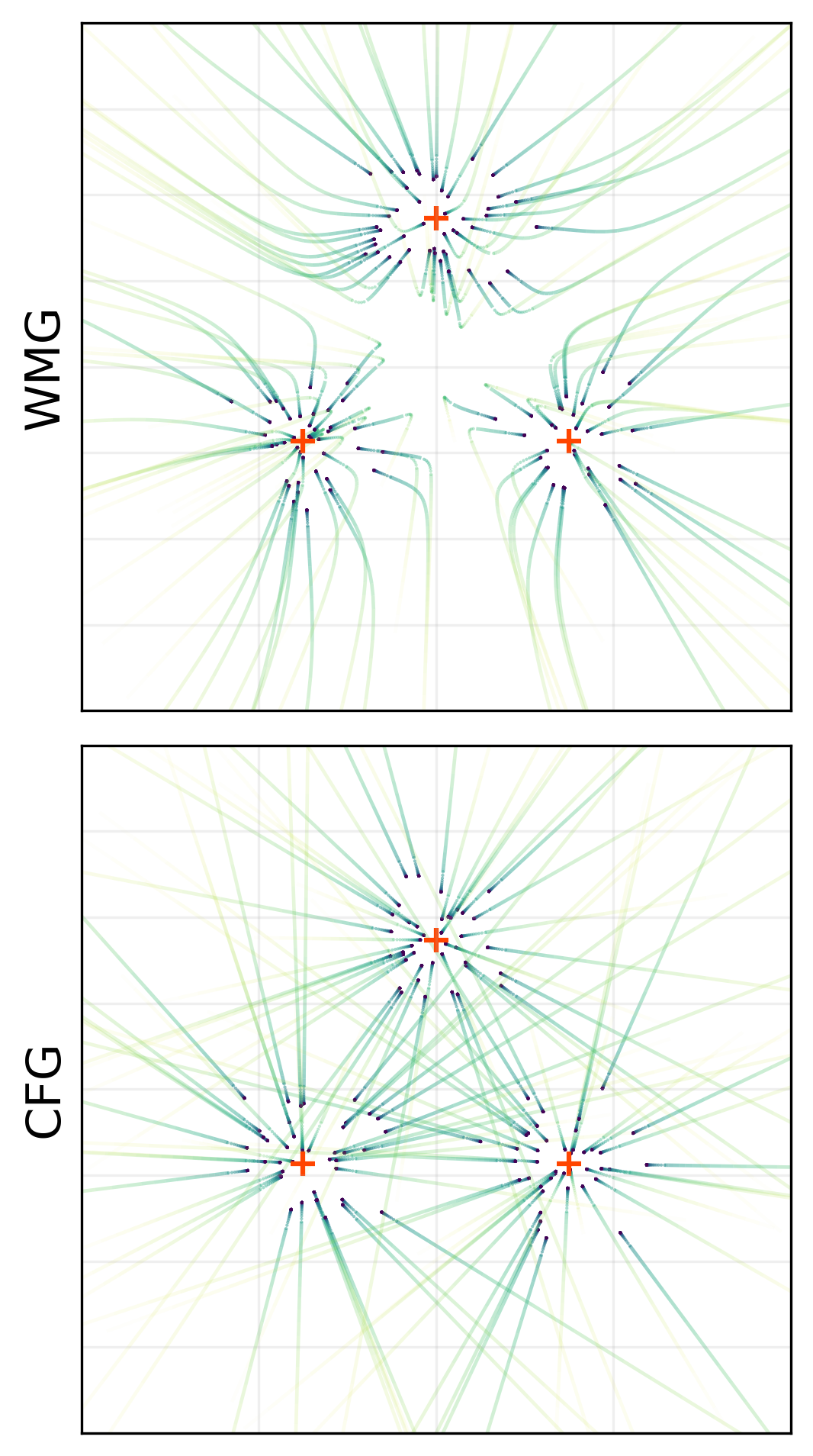}
                 \caption{$\w = 0$}
             \end{subfigure}
             \begin{subfigure}[b]{0.24\textwidth}
                 \includegraphics[scale=0.36]{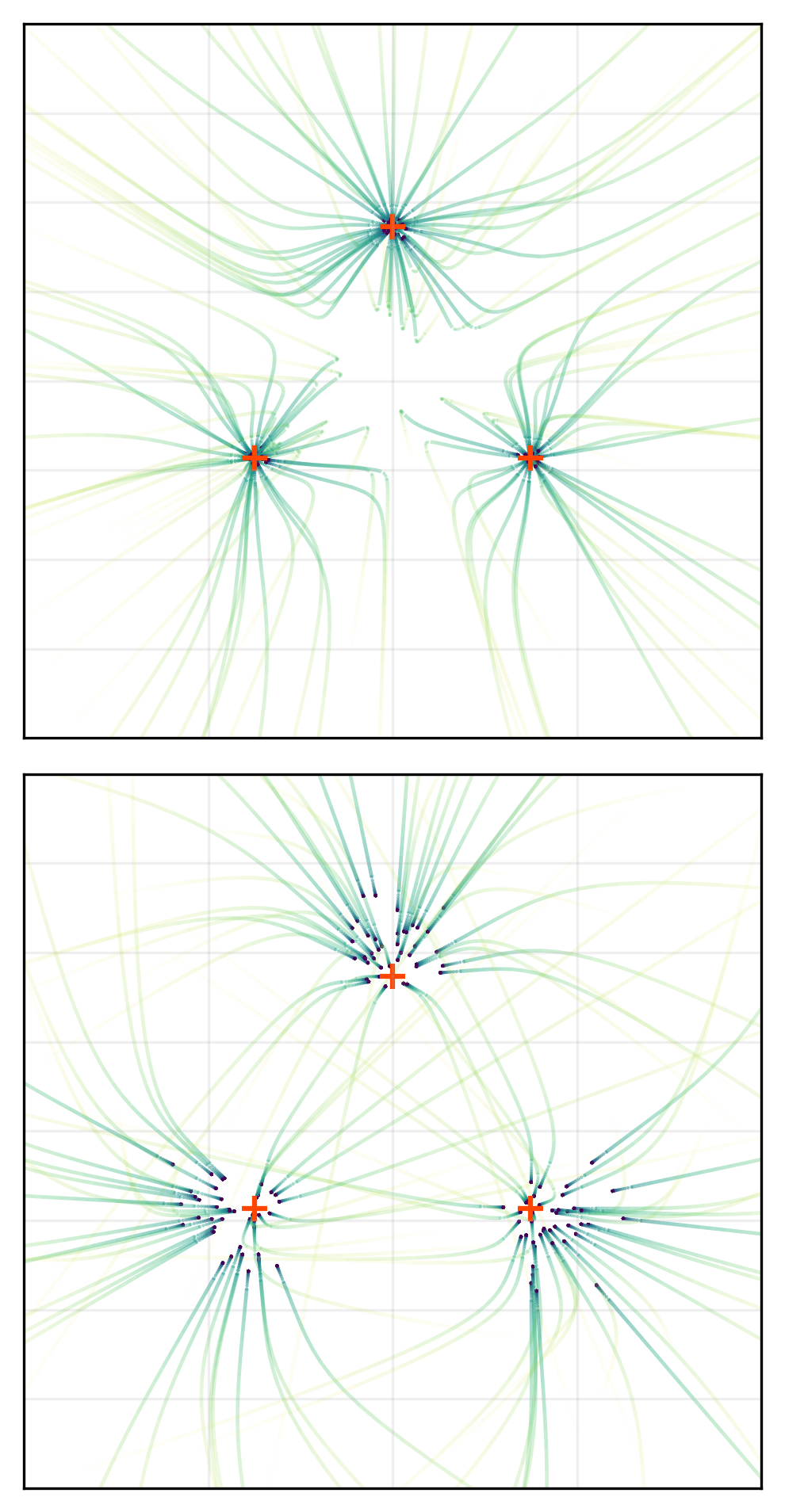}
                 \caption{$\w = w^*$}
             \end{subfigure}
             \begin{subfigure}[b]{0.24\textwidth}
                 \includegraphics[scale=0.36]{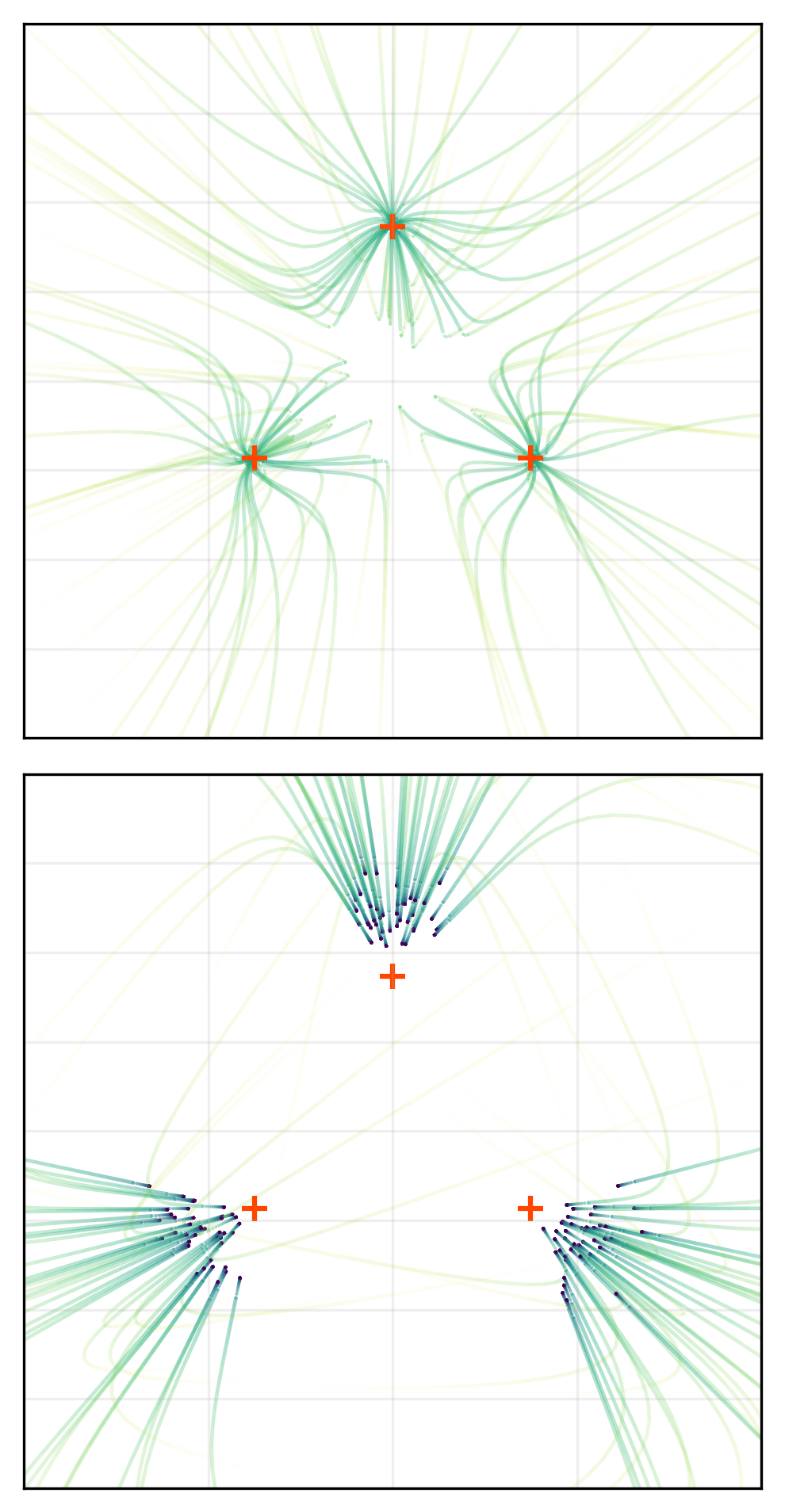}
                 \caption{$\w = 3w^*$}
             \end{subfigure}
             \begin{subfigure}[b]{0.24\textwidth}
                 \includegraphics[scale=0.36]{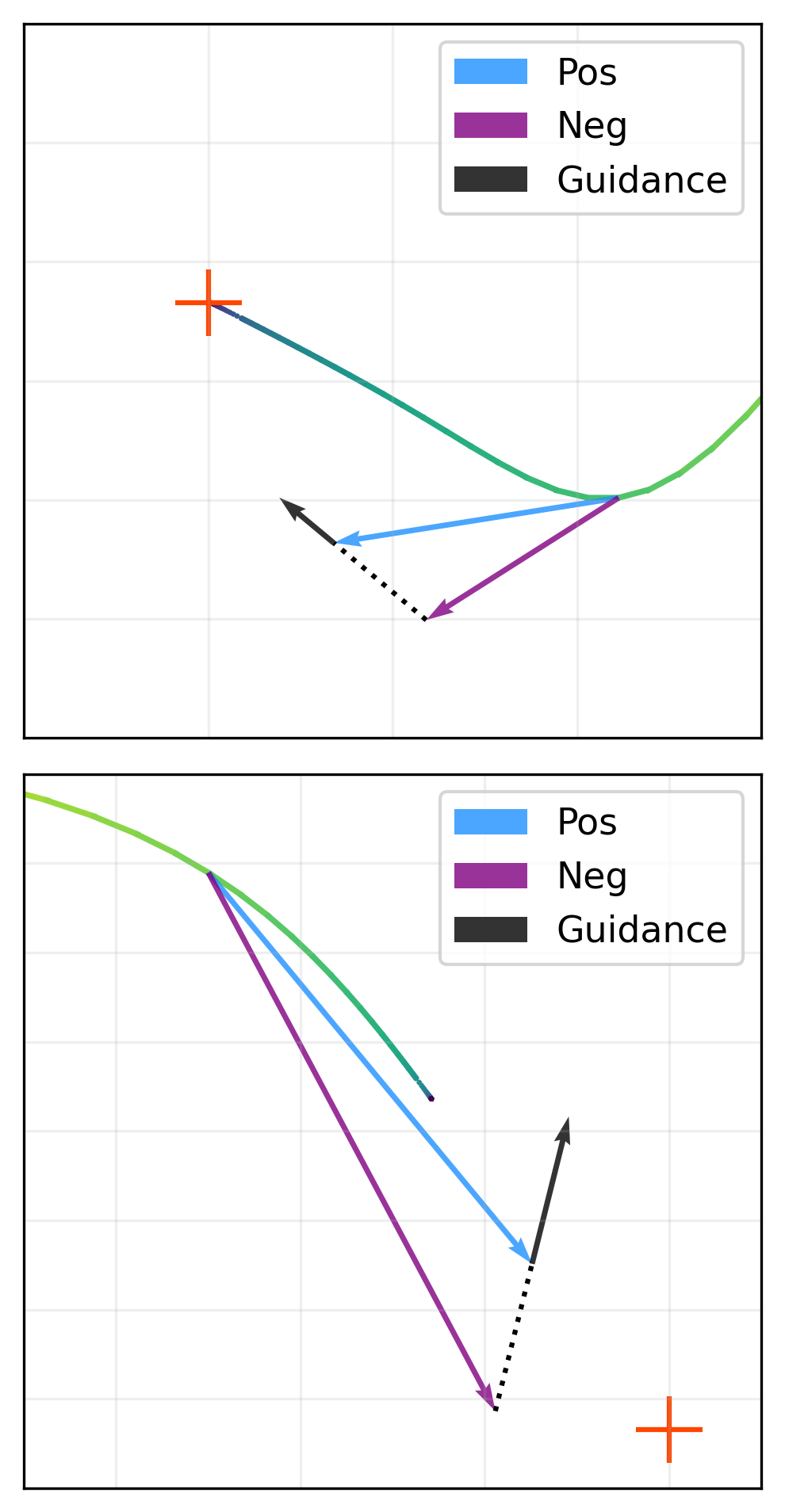}
                 \caption{Guidance triangle}
             \end{subfigure}
             };
        \node[rectangle,draw,minimum height=7mm,minimum width=7mm,opacity=0.8] (spypoint1) at (1.9,2.3){};
        \draw[opacity=0.8] (spypoint1) -- (3.45,2.3);  %
        \node[rectangle,draw,minimum height=7mm,minimum width=7mm,opacity=0.8] (spypoint2) at (1.45,-0.33){};
        \draw[opacity=0.8] (spypoint2) -- (3.45,-0.33);  %
    \end{tikzpicture}
    \caption{\textbf{Inference trajectories for our 2D-toy model. From left to right: (a)} $w=0$ yields the trajectories of the positive model. 
    \textbf{(b)} $w^*$ denotes the guidance weight that leads to best (CFG) or onset of saturating (\wmg) performance ($w^*=1$ for CFG and $w^*=5$ for \wmg). 
    \textbf{(c)} For large guidance weights, CFG exhibits large endpoint errors, while \wmg{} keeps small endpoint errors until it becomes unstable.
    \textbf{(d)} Enlarged views for single trajectories. Arrows show target prediction for a point on the trajectory. In this instance, the guidance correction (black arrow) pushes \wmg{} closer to the data point, unlike CFG.}
    \label{fig:toymodel}
    \vspace{-0.6cm}
\end{figure}

%% file: fig_wrap/swg_diagram.tex
\begin{wraptable}{r}{.5\textwidth}
\vspace{-0.2cm}
\centering
\includegraphics[width=\linewidth]{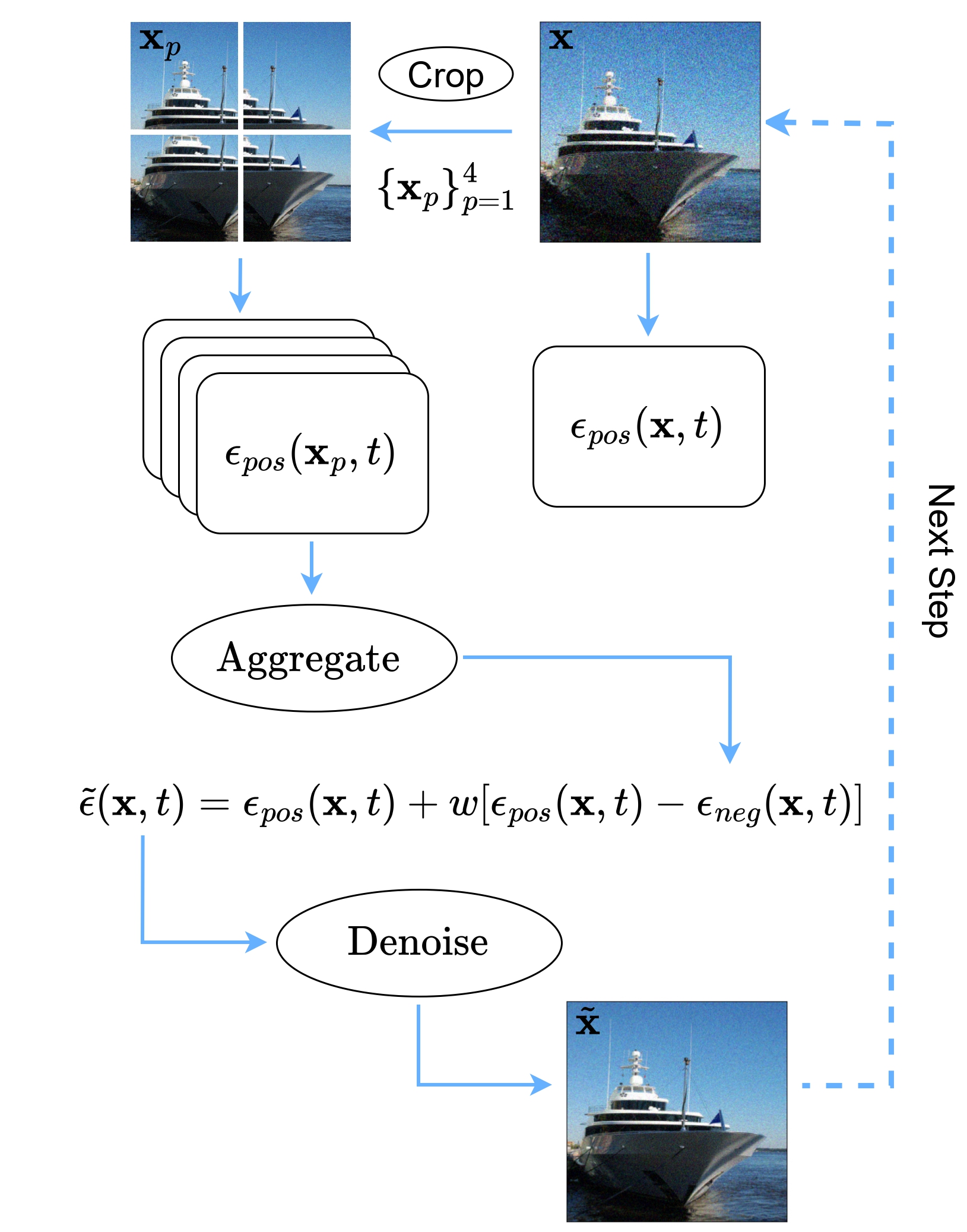}
\captionof{figure}{\textbf{An illustration of sliding window guidance (SWG)}. We use $\e_{\text{pos}}$ to independently process $N=4$ overlapping image crops. $\e_{neg}$ is generated by superimposing the processed crops and averaging the overlapping pixels.}
\label{fig:swg_diag}
\vspace{-0.9cm}
\end{wraptable}

%% file: merge/merge_1.tex
\begin{table}[t]
\begin{minipage}{.5\textwidth}
\begin{adjustbox}{width=\columnwidth}
                \begin{tabular}{cccc}
                \toprule
                Model  & Guidance & FID  & $\textnormal{FDD}$  \\
                \midrule
           \multirow{5}{*}{{\makecell{EDM2\\S-512}}} & \xmark  &  2.91 &  112  \\
             & $\text{RT}_{.8/1.2}$    & \textbf{1.79} & {46.5} \\ 
              & $\text{SWG}_{.1/.5}$    &  2.59 & 56.1 \\
                & $\text{M-SWG}_{.2/1.1}$    &  2.55 & 53.4 \\
                 & $\text{RT}_{.8/1.2}+\text{M-SWG}_{0/.25}$    &  1.79 & \textbf{40.9} \\
              \midrule
                   \multirow{4}{*}{{\makecell{EDM2\\XXL-512}}} & \xmark   &  2.29 & 49.7 \\ 
             & $\text{RT}_{.1/1.3}$    & \textbf{2.26} & 34.6 \\ 
                  &  $\text{SWG}_{.05/.2}$   & 2.61 & 37.7  \\
                   &  $\text{M-SWG}_{.05/.5}$   & 2.46 &  36.8 \\
                      & $\text{RT}_{.1/1.05}+\text{M-SWG}_{0/.1}$    &  2.26 & \textbf{32.9} \\
                \midrule
                   \multirow{3}{*}{{\makecell{DiT \\ XL-256} }}  & \xmark  & 9.83  & 213 \\  
                    &  $\text{SWG}_{.25/.5}$  & 4.00  & 80.6 \\
                       &  $\text{M-SWG}_{.5/1.5}$  & \textbf{3.30}  & \textbf{70.3} \\
        \midrule \midrule           
                \multicolumn{4}{l}{\textit{Unconditional generation}} \\
                   \multirow{4}{*}
                  {\makecell{EDM2\\S-512 }}  &  \xmark  &  13.7  & 250 \\
                  &  $\text{RT}_{1.6/2}$  &  \textbf{4.89} & 107.8 \\
                  &  $\text{SWG}_{.3/.5}$  &  10.2 & 167.4 \\
                   &  $\text{M-SWG}_{.5/.9}$  & 9.32 & 149.4 \\
                      & $\text{RT}_{1.6/1.35}+\text{M-SWG}_{0/.15}$    &  4.89 & \textbf{101.2} \\
                \bottomrule 
                \end{tabular} 
\end{adjustbox}
 
 \caption{{Training-free diffusion guidance sampling methods comparison.}} \label{tab:swg_results}
\end{minipage}
\hfill
\begin{minipage}{.47\textwidth}
\begin{adjustbox}{width=\columnwidth}
\begin{tabular}{ccc}
                \toprule
                Model  & Guidance & $\textnormal{FDD}$  \\
                \midrule
           \multirow{5}{*}{{\makecell{EDM2\\S-512}}} %
              &  {$\text{RCT}_{1.4}$}    &   { 42.1 }\\ 
              &   { $\text{RCT}_{1}+\text{SWG}_{.2}$}    &  39.2\\
               &   { $\text{RCT}_{1.2}+\text{M-SWG}_{.14}$}    &   \textbf{39.1} \\
              \cmidrule{2-3}
               &    CFG$^\dagger_{.8}$ & 52.9 \\
              &   { $\text{CFG}^{\dagger}_{.3}+\text{M-SWG}_{.3}$}    &   \textbf{51.9} \\
              \midrule
                   \multirow{4}{*}{{\makecell{EDM2\\XXL-512}}} %
                   &  { $\text{RCT}_{1}$}   &  30.3  \\
                   &  { $\text{RCT}_{.9}+\text{M-SWG}_{.05}$}   &  \textbf{29.8} \\
                   \cmidrule{2-3}
               &    CFG$^{\dagger}_{.4}$ & 36.7 \\
              &   { $\text{CFG}^{\dagger}_{.2}+\text{M-SWG}_{.05}$}    &   {\textbf{36.2}} \\
                \midrule
                   \multirow{3}{*}{{\makecell{DiT\\XL-256} }}  %
                   &  {  $\text{CFG}_{2.05}$}   &   { 54.3} \\
                   &  {  $\text{CFG}_{.7}+\text{SWG}_{.1}$}  &   50.6  \\
                    &  {  $\text{CFG}_{.6}+\text{M-SWG}_{.5}$}  &    {\textbf{49.1}} \\
\midrule \midrule
                \multicolumn{3}{l}{\textit{Unconditional generation}} \\
                   \multirow{2}{*}
                  {\makecell{EDM2\\S-512 }}  %
                  &  $\text{RCT}_{1.8}$  &  97.9 \\
                   &   $\text{RCT}_{1.3} + \text{M-SWG}_{.15}$  & {\textbf{91.8}} \\
                \bottomrule 
                \end{tabular}
                \end{adjustbox}
                \caption{M-SWG enhances existing state-of-the-art methods. The dagger $\dagger$ indicates that a smaller capacity model is used.} \label{tab:combine_swg_cfg_rct_iccv}
\end{minipage}
\caption*{\textbf{Tables\ \ref{tab:swg_results}-\ref{tab:combine_swg_cfg_rct_iccv}: We report \fdd{} ($\downarrow$) and FID ($\downarrow$) on ImageNet for several benchmarks.} The subscripts indicate the chosen guidance scale that yields the optimal guidance weights for FID/FDD based on a hyperparameter search.}
\end{table}

%% file: fig_wrap/pareto_study.tex
\begin{figure}[t]
\centering\includegraphics[width=\linewidth]{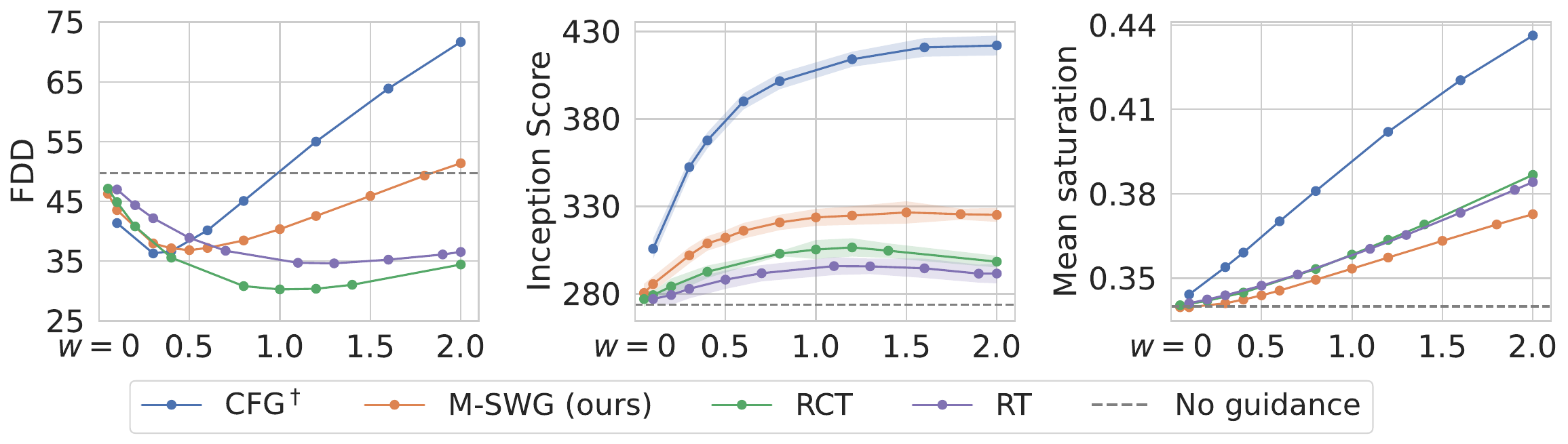}
\vspace{-0.5cm}
    \caption{\textbf{Visualizing the achieved tradeoff by varying the guidance scale $w$ (x-axis) for FDD ($\downarrow$, left), Inception score (IS $\uparrow$, center), and saturation ($\downarrow$, right) using EDM2-XXL}. M-SWG achieves superior perceptual quality as measured by the IS compared to RT and RCT, while not exhibiting undesired oversaturation as CFG$^\dagger$, on average.}
    \vspace{-0.5cm}
    \label{fig:pareto}
\end{figure}

%% file: merge/merge_2_tables.tex
\begin{table}[t]
\begin{minipage}{.3\textwidth}
\begin{adjustbox}{width=\columnwidth}
\begin{tabular}{lcc}
                 \toprule
            N, k, r & FDD & FID \\
            \hline
            \xmark &  112.2  &  2.92 \\
            \midrule
            4, 32, 0 &  ${55.7}_{0.3}$ & $2.72_{0.1}$ \\
            4, 40, 0.4 & $56.1_{0.5}$  & $2.59_{0.1}$ \\  
            4, 48, 0.6 & $60.7_{0.5}$  & $\mathbf{2.56}_{0.1}$ \\ \hline
            9, 32, 0.5  & $\mathbf{55.4}_{0.3}$  & $2.69_{0.1}$ \\
            9, 24, 0.63  & $58.1_{0.2}$  & $2.89_{0.1}$  \\
            16,16,0  & $64.0_{0.2}$ &  $3.15_{0.1}$  \\
           \bottomrule
        \end{tabular}
\end{adjustbox}   \caption{{Varying kernel size $k$ and number of crops $N$.}}
  \label{tab:ablation-kernel}
\end{minipage}
\hfill
\begin{minipage}{.31\textwidth}
\begin{adjustbox}{width=\columnwidth}
 \begin{tabular}{lcc}
            \toprule
            Guidance &  Interval  &  FDD   \\ 
            \midrule
           \xmark                    &  -  & $112.2$       \\ 
            \midrule

            RT$_{1.2}$                   & -          & $46.5$    \\ 
             + interval$_{1.8}$            & 10-20         & \textbf{43.8}  \\ 
             \hline
            SWG$_{0.5}$                     & -          & $56.1$   \\ 
             + interval$_{1.2}$            & 15-20 & $51.3$    \\ 
        \hline 
        M-SWG$_{1.1}$                     & -          & 53.4   \\ 
             + interval$_{1.7}$            & 13-23 & 51.7   \\ 
            \bottomrule
        \end{tabular}
                \end{adjustbox}
                    \caption{Applying guidance only at an interval (32 steps).}  \label{tab:ablation-interval}
\end{minipage}
\hfill
\begin{minipage}{.35\textwidth}
    \begin{adjustbox}{width=\columnwidth}
        \begin{tabular}{lcc}
    \toprule
    & \fdd & FID \\
    \midrule
   CFG$^\dagger$ \cite{ho2022classifierfreediffusionguidance} & $52.9_{0.8}$ & $2.44_{0.1}$  \\
    \midrule
     \multicolumn{3}{l}{\textit{\wmg{} methods}}\\
    a) Reduced training & $46.5_{1.2}$ & $1.79_{0.8}$  \\
    b) Reduced capacity & ${70.0}_{1.4}$ & $2.20_{0.7}$  \\
    c) RCT (a + b)  & $\mathbf{42.1_{1.4}}$ & $\mathbf{1.67}_{0.8}$  \\
    \hline
    d) WD fine-tuning & $52.1_{0.6}$ & $2.46_{0.1}$  \\
    e) WD + (a) & $43.9_{0.9}$ & $2.25_{0.2}$  \\
    \bottomrule
\end{tabular}
    \end{adjustbox}
       \caption{A comparison of WMG strategies, including the novel weight decay (WD) variants.}     \label{tab:verification_edm_wmg_in1k}
\end{minipage}
 \caption*{\textbf{Tables \ref{tab:ablation-kernel}-\ref{tab:verification_edm_wmg_in1k}: Additional experiments on ImageNet-512 using EDM2-S.} The subscripts indicate the chosen guidance scales based on a hyperparameter search.}
\vspace{-0.1cm}
\end{table}

%% file: figs/oranges.tex
\begin{figure}
    \centering
   \includegraphics[width=0.95\textwidth]{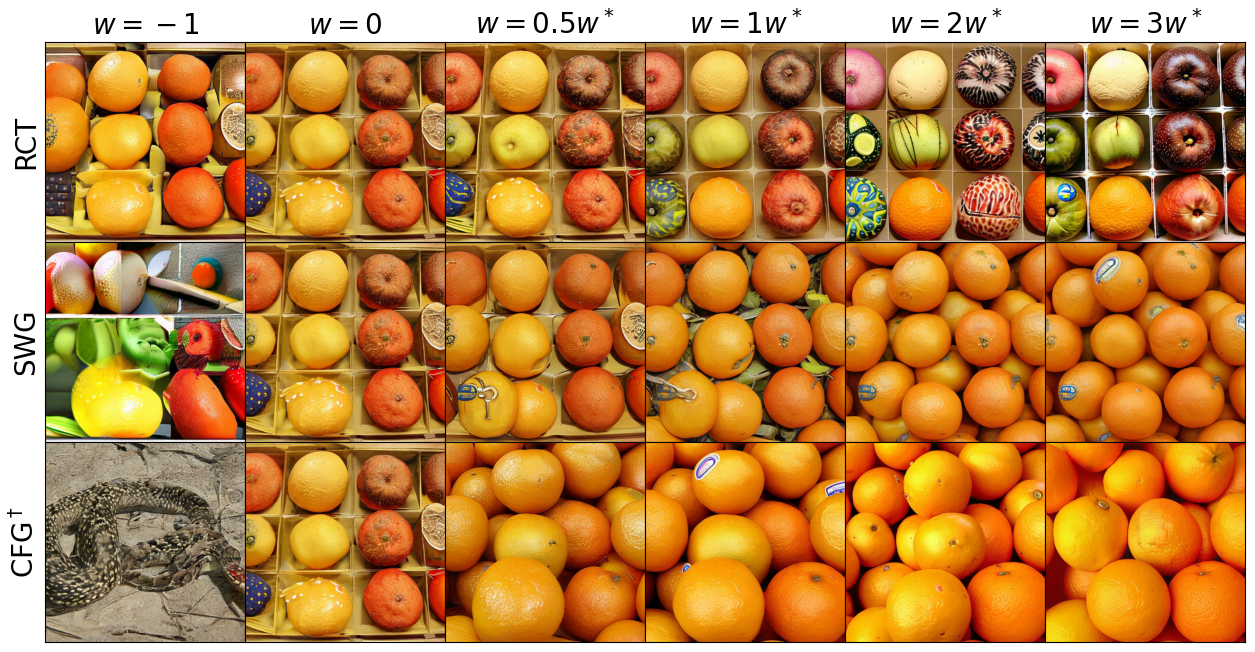}
    \caption{\textbf{Samples generated by different guidance methods (rows) with different guidance weights (columns) but the same initial noise and condition (orange class) using EDM2-XXL.} Guidance weights $w$ are scaled by $w^*$, the individual weight for each method corresponding to the lowest \fdd. The first column shows samples using $\e_{neg}$, which corresponds to setting $w=-1$ in Eq.~(\ref{eq:wmg}).}
\label{fig:oranges}
\vspace{-0.4cm}
\end{figure}

%% file: appendix.tex
\setlength{\tabcolsep}{6pt} %
\renewcommand{\arraystretch}{1} %

\section{Additional implementation details and experiments}
\textbf{Constrast.} In \Cref{fig:contrast}, similar to saturation, we show that the discrepancy between M-SWG and other state-of-the-art methods is significant. The combination of competitive IS and FDD, with significantly less oversaturation and almost no overcontrast on average, suggests that samples remain relatively closer to the training data distribution. Contrast is computed as the root mean square contrast of the samples, meaning the standard deviation of intensities after greyscale conversion, similar to \cite{peli1990contrast,sadat2024eliminating}. 

\input{fig_wrap/wrap_contrast}

\textbf{DiT: guidance scales w.r.t. Frechet distances.} Similar to \cite{kaplan2020scaling}, we have assumed that the generative error of $\e_{\text{pos}}$ can be primarily attributed to scaling factors, such as training time and number of parameters. In the main manuscript, we illustrate that the largest best-performing publicly available Unet architecture EDM2-XXL shows no improvement w.r.t. FID for SWG and only minimal improvement for reduced training (RT). However, this is not in line with DiT-XL/2. In \Cref{fig:pareto_guidance_scales}, we demonstrate that the unguided generation can be significantly improved using SWG w.r.t. to both FID and FDD. We suspect that the observed discrepancy can be attributed to the significantly better inductive bias of EDM2-XXL, which leaves less room for improvement using guidance.
\input{fig_wrap/guidance_scales_supp}

\textbf{SWG implementation details and pseudocode.} To make SWG compatible with DiT, we add positional encoding interpolation on DiT-XL to process various resolutions. The choice of $k\approx \frac{5}{8} H$ is determined by practical considerations of Unets. Specifically, $k$ must satisfy $\frac{k}{2^n} \in \mathbb{N}$, where $n$ is the number of downsampling steps of the Unet, and the diffusion model $\e_{\text{pos}}$ must be able to process multiple image resolutions. %

We provide an implementation of SWG in the \Cref{pseudo-code-swg}. 
\input{figs/swg-pseudo}

\textbf{Time overhead at sampling time.} In \Cref{tab:benchmark_time}, we present the time needed to produce 512 samples for various models and guidance methods and their corresponding estimate in images per second per GPU. We emphasize that we have naively implemented SWG, as shown in the pseudocode, and further optimization can help bridge the time overhead.

\textbf{Guidance interpolation.} \Cref{fig:interpolation,fig:4-interp} provide interpolations of guidance methods following 
\begin{align} \label{eq:interpolation}
    \tilde \e(\x, t) 
    = &\e_{\text{pos}}(\x,t) + \sum_i \alpha_i \w_i [\e_{\text{pos}}(\x,t) - \e^{(i)}_{\text{neg}}(\x,t)]
\end{align}
where $\w_i$ are the optimized guidance scales w.r.t. FDD for each method and $\alpha_i$ are set to form a convex combination between guidance terms. 

\input{figs/interpolation}
\input{figs/4-interpolation}

\input{tables/benchmark} %

\input{tables/sota_edm}

\input{tables/sota_edm_settings}

\clearpage
\input{toy_model_details}
\clearpage

\input{fig_wrap/rct_swg_supp}
\input{fig_wrap/rct_swg_guidance_scales}
\input{fig_wrap/teaser_supp}

%% file: fig_wrap/wrap_contrast.tex
\begin{figure}[th]
\centering\includegraphics[width=\linewidth]{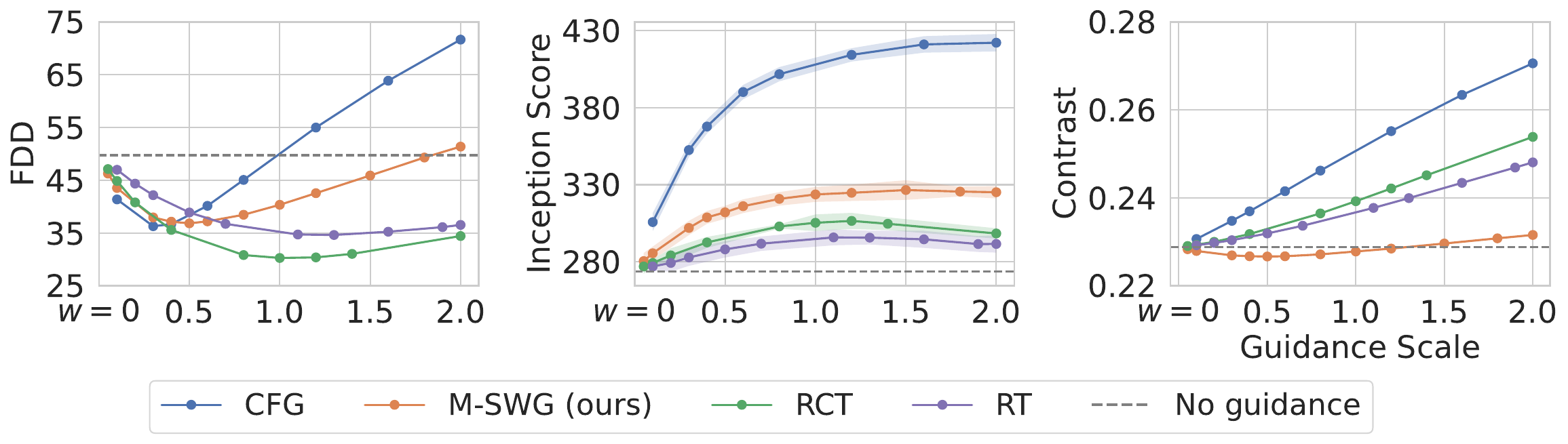}
    \caption{\textbf{Visualizing the achieved tradeoff by varying the guidance scale $w$ (x-axis) for FDD  ($\downarrow$, left), Inception score (IS $\uparrow$, center), and contrast ($\downarrow$, right) using EDM2-XXL}. M-SWG does \textbf{not} exhibit undesired overcontrast compared to other state-of-the-art methods.}
    \vspace{-0.1cm}
    \label{fig:contrast}
\end{figure}

%% file: fig_wrap/guidance_scales_supp.tex
\begin{figure}[bht]
\centering
        \includegraphics[width=0.75\textwidth]{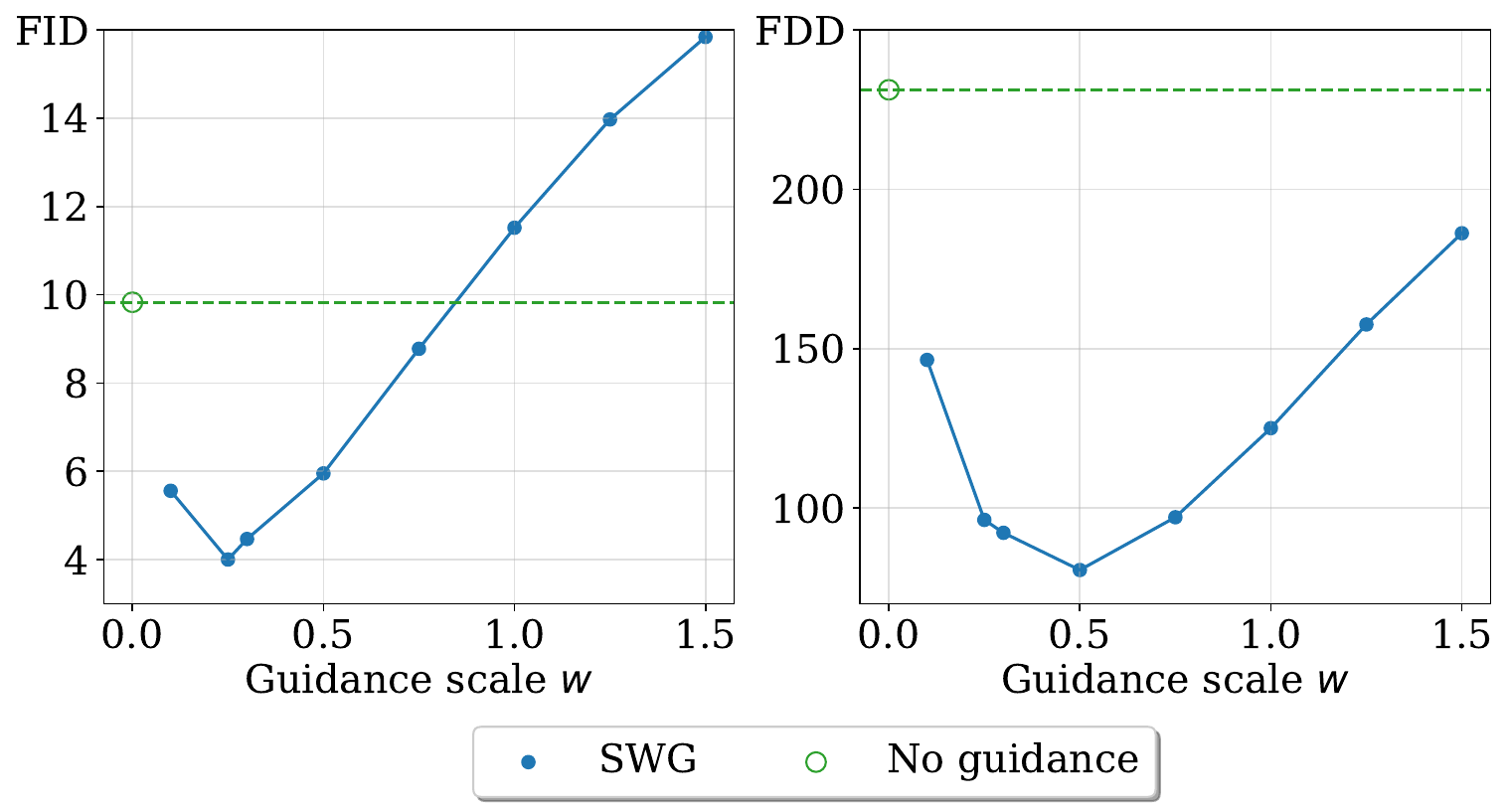}  \label{fig:dit-xl-guidance-curve} 
    \caption{DiT-XL/2 on ImageNet 256x256.Frechet distances (\textit{y-axis}) versus guidance scales (\textit{x-axis}) using InceptionNetv3 (\textbf{FID, left}) and DINOv2 ViT-L (\textbf{FDD, right}).}
    \label{fig:pareto_guidance_scales}
\end{figure}

%% file: figs/swg-pseudo.tex
\newcommand{\pluseq}{\mathrel{+}=}

\definecolor{codeblue}{rgb}{0.1, 0.1, 0.8}
\definecolor{codegreen}{rgb}{0.1, 0.6, 0.1}
\definecolor{codegray}{rgb}{0.5, 0.5, 0.5}
\definecolor{codepurple}{rgb}{0.6, 0.1, 0.6}
\definecolor{backcolour}{rgb}{0.95, 0.95, 0.92}

\lstdefinestyle{mystyle}{
    backgroundcolor=\color{backcolour},   
    commentstyle=\color{codegreen},
    keywordstyle=\color{codeblue},
    numberstyle=\tiny\color{codegray},
    stringstyle=\color{codepurple},
    basicstyle=\ttfamily\footnotesize,
    breakatwhitespace=false,         
    breaklines=true,                 
    captionpos=b,                    
    keepspaces=true,                 
    numbers=left,                    
    numbersep=5pt,                  
    showspaces=false,                
    showstringspaces=false,
    showtabs=false,                  
    tabsize=2
}

\lstset{style=mystyle}
    \begin{figure}[h]
      \centering
    \begin{minipage}{0.75\linewidth}
        \begin{lstlisting}[language=Python]
import torch

def SWG(x, e_pos, t, labels, N, k, w):
  # x: noisy images of size [bs, c, H, H] 
  # e_pos: network i.e. Unet
  # t: timestep
  # labels: class conditions 
  # N: number of crops in [4,9,16,...]
  # k: crop size
  # w: guidance scale
  
  steps_per_dim = int(torch.sqrt(N))
  y_pos = e_pos(x, t, labels)
  bs, c, H, _ = x.shape
  y_neg = torch.zeros_like(x)
  overlap = torch.zeros_like(x)
  stride = (H - k) // (steps_per_dim - 1)
  for i in range(steps_per_dim):
    for j in range(steps_per_dim):
      # left edge (le), right edge (re)
      le, re = stride * i, stride * i + k  
      # top edge (te), bottom edge (be)
      te, be = stride * j, stride * j + k  
      # x_p is of shape [bc, c, k, k]
      x_p = x[:, :, te:be, le:re] 
      y_crop = e_pos(x_p, t, labels)
      y_neg[:, :, te:be, le:re] += y_crop
      o_p = torch.ones_like(y_crop)
      overlap[:, :, te:be, le:re] += o_p
  # pixel averaging
  y_neg = y_neg / overlap
  # Equation 2 in the main text
  return y_pos + w * (y_pos - y_neg)   
        \end{lstlisting} 
    \label{pseudo-code-swg}
    \end{minipage}
    \caption{Example pseudocode for sliding window guidance (SWG) in Python using the torch library.}
\end{figure}

%% file: figs/interpolation.tex
\begin{figure*}
    \centering
    \includegraphics[width=1\textwidth]{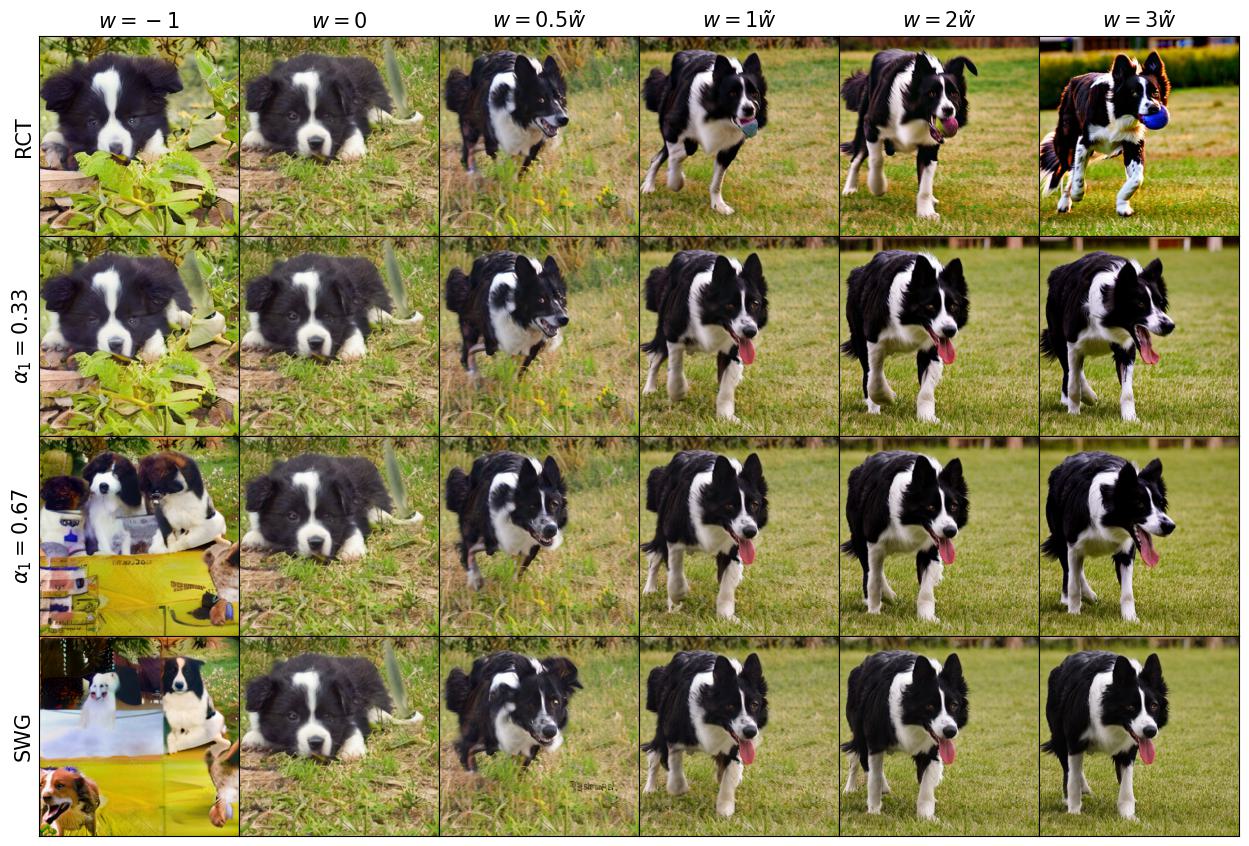}
    \caption{Samples generated by interpolating RCT and SWG with different guidance weights, but the same initial noise. The noise predictor $\e_{pos}$ is based on the EDM2-XXL network trained on ImageNet-512. \textbf{Columns from left to right:} guidance weights $w$ are scaled by $\tilde w$, with $\tilde w$ the individual weight for each method corresponding to the lowest \fdd{}  ($\tilde w=1$ for RCT and $\tilde w=0.2$ for SWG). The first column shows samples using $\e_{neg}$ as a noise predictor ($w=-1$). Interpolations are computed according to \Cref{eq:interpolation} with a convex combination of the guidance terms parametrized by $\alpha_1$ and $\alpha_2 = 1-\alpha_1$.}
    \label{fig:interpolation}
\end{figure*}

%% file: figs/4-interpolation.tex
\begin{figure*}[h]
    \centering
    \begin{subfigure}[b]{0.49\textwidth}
        \includegraphics[width=1\textwidth]{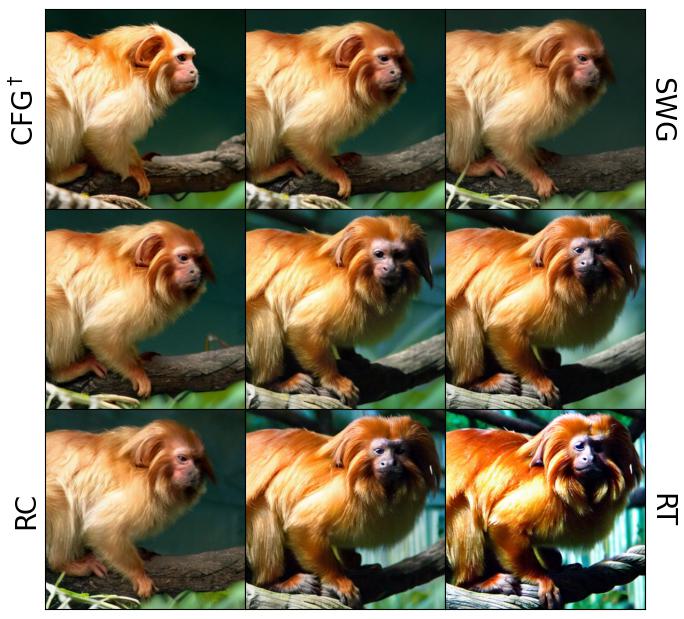}
    \end{subfigure}
    \hfill
    \begin{subfigure}[b]{0.49\textwidth}
        \includegraphics[width=1\textwidth]{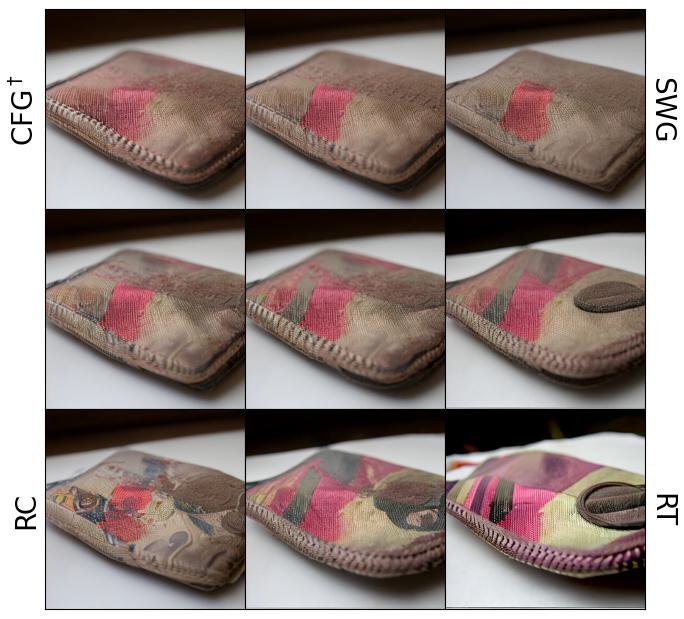}
    \end{subfigure}
    \par\bigskip %
    \begin{subfigure}[b]{0.49\textwidth}
        \includegraphics[width=1\textwidth]{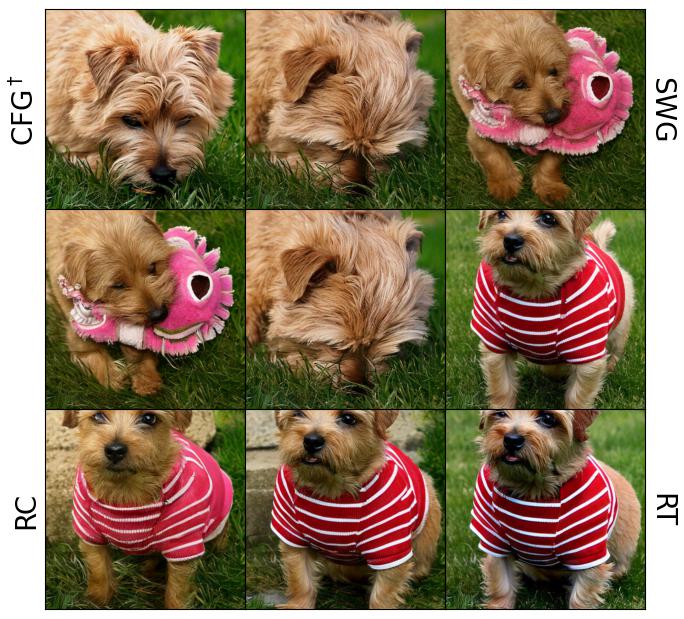}
    \end{subfigure}
    \hfill
    \begin{subfigure}[b]{0.49\textwidth}
        \includegraphics[width=1\textwidth]{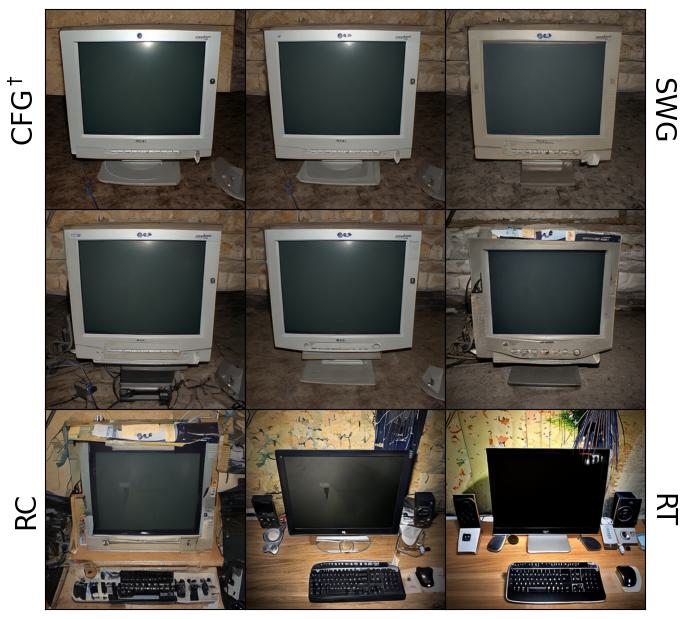}
    \end{subfigure}
    \caption{Samples generated by interpolating CFG, SWG, RC (reduced capacity), and RT (reduced training) with different guidance weights but the same initial noise. The noise predictor $\e_{pos}$ is based on the EDM2-XXL network trained on ImageNet-512. Respective weights are CFG: $w=0.6$, SWG: $w=0.2$, RT: $w=1.0$ and RC: $w=1.0$. Interpolations are computed according to \Cref{eq:interpolation}, such that the sum over guidance terms is a convex combination, with $\alpha_i = 1$ in the corners and linear interpolation in between.%
    }
    \label{fig:4-interp}
\end{figure*}

%% file: tables/benchmark.tex
\begin{table*}
    \centering

    \begin{tabular}{cccccc}
        \toprule
        Model/Res  & Guidance & \makecell{VRAM (GB) \\ per GPU }  & \makecell{Batch \\ size}  & \makecell{Time for \\ 512 samples \\ in minutes ($\downarrow$)} & \makecell{Estimate of \\ Imgs/s per \\ GPU ($\uparrow$)}  \\
        \midrule
        \multirow{6}{*}{{\makecell{EDM2-S \\ 512$\times$512}}} & \xmark  &   15.2 & 256 &   1.36 & 12.54 \\
        & $\text{SWG}$ (r=0.4)    & 15.2   & 256  & 2.10 & 8.12  \\
        & $\text{SWG}$ (r=0)    & 15.2  &  256  & 1.85 & 9.22  \\
        &    $\text{RT}$    &   15.2 &       256  & 1.77 & 9.64  \\
        &    $\text{RCT}$    &  15.2  &     256  & 1.61 & 10.67   \\
        &    $\text{CFG}^{\dagger}$ & 15.2 & 256  & 1.61 & 10.67        \\ 
        \hline \hline
        \multirow{3}{*}{{\makecell{DiT-XL \\ 256$\times$256 }}} & \xmark & 17.9  & 32  & 4.61 &  3.70  \\
        &    $\text{CFG}$    & 17.9 & 32  & 8.45 & 2.02      \\ 
         & $\text{SWG}$    &  17.9  & 32  & 11.37 &  1.50  \\
        \bottomrule 
    \end{tabular} \label{tab:benchmark_time}
        \caption{All benchmarks were conducted using 2 Nvidia A100 GPUs with 40GB of VRAM. We benchmark sliding window guidance (\textit{SWG}), classifier-free guidance (\textit{CFG}), classifier-free guidance using a smaller capacity unconditional model (\textit{CFG}$^\dagger$), reduced training (\textit{RT}), and reduced capacity training (\textit{RCT}). The symbol \xmark \ stands for \textit{No guidance} ($w=0$). Crucially, SWG \emph{is implemented in a naive way} with $N=4$ (as shown in the provided pseudocode) and can be further optimized, which is left to future work. The scalar $r$ denotes the overlap ratio per dimension, where $r=0$ refers to no overlap.}
\end{table*}

%% file: tables/sota_edm.tex
\begin{table}
    \centering
    \small
    \begin{tabular}{lccccc}
        \toprule
         & \multicolumn{3}{c}{{EDM}} & \multicolumn{2}{c}{{EDM2-S}}  \\
        \cmidrule(lr){2-4} \cmidrule(lr){5-6} 
         & CIFAR-10 & CIFAR-100 & FFHQ-64 & \multicolumn{2}{c}{ImageNet-512} \\
        \cmidrule(lr){2-2} \cmidrule(lr){3-3} \cmidrule(lr){4-4} \cmidrule(lr){5-6} 
         & \fdd & \fdd & \fdd & \fdd & FID \\
        \midrule
        No Guidance  & $139.4$ & $196.6$   & $174.8$ & $112.2$ & $2.92$  \\
        \midrule
        CFG \cite{ho2022classifierfreediffusionguidance}    & $99.9_{0.6}$  & $126.6_{0.6}$  &  $- $  & $52.9_{0.8}^\dagger$ & $2.44_{0.1}^\dagger$  \\
        \midrule
        \textit{\wmg{} methods} \\
        a) Reduced training                     & $60.8_{1.1}$             &  $81.2_{1.3}$             &  $94.0_{1.0}$             &  $46.5_{1.2}$  & $1.79_{0.8}$  \\
        b) Reduced capacity                     & $62.4_{0.9}$             &  $93.8_{1.1}$             &  $102.5_{1.3}$            &  $70.0_{1.4}$  & $2.20_{0.7}$  \\
        c) RCT (a + b)                          & $62.1_{0.8}$             &  $87.0_{1.0}$             &  $95.3_{1.0}$             &  $\mathbf{42.1_{1.4}}$  & $\mathbf{1.67}_{0.8}$  \\
        d) Weight decay fine-tuning             & $65.0_{1.3}$             &  $88.1_{1.4}$             &  $111.2_{2.4}$            &  $52.1_{0.6}$  & $2.46_{0.1}$  \\
        e) Weight decay + (a)                   & $\mathbf{59.0_{0.9}}$    &  $\mathbf{75.9_{1.1}}$    &  $\mathbf{92.6_{1.0}}$    &  $43.9_{0.9}$           & $2.25_{0.2}$  \\
        f) Weight decay re-training             & $64.7_{1.3}$             &  $93.3_{1.3}$             &  $133.0_{1.4}$            &  $-$           & $-$  \\
        \bottomrule
    \end{tabular}
    \caption{Extension of Table 5 (main text) with EDM \cite{karras_edm}. FDD ($\downarrow$) and {FID ($\downarrow$)}, with the guidance weight in subscript. FFHQ-64 is the only unconditional task; all other experiments used class labels as model conditions for both networks. \textsuperscript{\textdagger} The negative model additionally has reduced capacity (EDM2-XS).}
    \label{app:sota_edm}
\end{table}

%% file: tables/sota_edm_settings.tex
\begin{table}
    \centering
    \small
    \begin{tabular}{|l|c|c|c|c||c|c|c|c|c|}
        \hline
        & \multicolumn{8}{c|}{EDM2-S, ImageNet-512} \\
        \cline{2-9}
         & \multicolumn{4}{c||}{FDD} & \multicolumn{4}{c|}{FID} \\
        \hline
        Settings & c$_{\text{pos}}$ & c$_{\text{neg}}$ & Size & $\lambda$ & c$_{\text{pos}}$ & c$_{\text{neg}}$ & Size & $\lambda$ \\
        \hline
        No guidance & 2147 & 2147 & S & 0 & 2147 & 2147 & S & 0 \\ 
        CFG$^{\dagger}$ & 2147 & 2147 & S & 0 & 2147 & 2147 & S & 0 \\ 
        \hline 
        \textit{\wmg{} methods} & & & & & & & & \\
        a) Reduced training                    & 2147  & 67    & S     & 0     & 2147  & 134   & S     & 0 \\
        b) Reduced capacity                    & 2147  & 2147  & XS    & 0     & 2147  & 2147  & XS    & 0 \\
        c) RCT (a + b)                         & 2147  & 134   & XS    & 0     & 2147  & 268   & XS    & 0 \\
        d) Weight decay fine-tuning            & 2147  & 2233  & S     & 2e-5  & 2147  & 2243  & S     & 2e-5 \\
        e) Weight decay + (a)                  & 2147  & 63    & S     & 2e-5  & 2147  & 63    & S     & 2e-5 \\
        \hline
    \end{tabular}
    \caption{Settings for Table 5 (main text) on ImageNet-512 using EDM2-S. \textbf{Columns} specify the checkpoint of the positive model c$_{\text{pos}}$ in million images seen, the checkpoint of the negative model c$_{\text{neg}}$ in million images seen, the model size (EDM2 variants), and the weight decay parameter $\lambda$.}
    \label{app:settings-edm2}
\end{table}

{
\setlength{\tabcolsep}{2.5pt}           %
\begin{table}
    \centering
    \small
    \begin{tabular}{|l|c|c|c|c||c|c|c|c||c|c|c|c|}
        \hline 
        & \multicolumn{12}{c|}{EDM} \\
        \cline{2-5} \cline{6-9} \cline{10-13} 
        & \multicolumn{4}{c||}{CIFAR-10} & \multicolumn{4}{c||}{CIFAR-100} & \multicolumn{4}{c|}{FFHQ-64} \\
        & \multicolumn{4}{c||}{FDD} & \multicolumn{4}{c||}{FDD} & \multicolumn{4}{c|}{FDD} \\
        \hline
        Settings & c$_{\text{pos}}$ & c$_{\text{neg}}$ & Size & $\lambda$ & c$_{\text{pos}}$ & c$_{\text{neg}}$ & Size & $\lambda$ & c$_{\text{pos}}$ & c$_{\text{neg}}$ & Size & $\lambda$ \\
        \hline
        No guidance & 500 & 500 & F & 0 &  500 & 500 & F & 0 &  500 & 500 & F & 0 \\ 
        CFG         & 500 & 500 & F & 0 &  500 & 500 & F & 0 &  500 & 500 & F & 0 \\ 
        \hline
        \textit{\wmg{} methods} & & & & & & & & & & & & \\
        a) Reduced training         & 500 & 20  & F & 0      &  500 & 20  & F & 0       &  500 & 20  & F & 0 \\
        b) Reduced capacity         & 500 & 500 & S & 0      &  500 & 500 & S & 0       &  500 & 500 & S & 0 \\
        c) RCT (a + b)              & 500 & 200 & S & 0      &  500 & 100 & S & 0       &  500 & 100 & S & 0 \\
        d) WD fine-tuning           & 500 & 560 & F & 5e-4   &  500 & 520 & F & 5e-4    &  500 & 540 & F & 2e-3 \\
        e) WD and reduced training  & 500 & 30  & F & 5e-4   &  500 & 20  & F & 5e-4    &  500 & 20   & F & 2e-3 \\
        f) WD re-training           & 500 & 500 & F & 5e-4   &  500 & 500 & F & 5e-4    &  500 & 500 & F & 2e-3 \\
        \hline 
    \end{tabular}
    \caption{Settings for \Cref{app:sota_edm} on small-scale benchmarks using EDM. On CIFAR-10, CIFAR-100, and FFHQ-64. CFG uses an unconditional but otherwise identical model to the positive model as a negative model. \textbf{Columns} specify the checkpoint of the positive model c$_{\text{pos}}$ in million images seen, the checkpoint of the negative model c$_{\text{neg}}$ in million images seen, the model size (F for full and S for small), and the weight decay parameter $\lambda$. The full model corresponds to the default architecture hyperparameter settings as detailed in \cite{karras_edm}. The small model has a reduced number of resolution levels and channel size (cres=1,1 for CIFAR-10 / CIFAR-100 and cres=1,1,1 for FFHQ) and a reduced number of blocks per level (num\_blocks=2 for CIFAR-10 / CIFAR-100). %
    }
    \label{app:settings-edm}
\end{table}
}

%% file: toy_model_details.tex
\section{Toy example details} \label{app:toy}
This section presents the derivation of the toy example from the main text. The code for computing the trajectories will be released and allows for the visualization of guidance effects for different combinations of $\e_{\text{pos}}$, $\e_{\text{neg}}$, and datasets. We introduce with time $t\in[0,1]$ a variable that denotes the progress of denoising, which starts from a normally distributed sample, $\x(1)$, and ends at a fully denoised sample, $\x(0)$. In the following, we denote by $\y$ a sample from the data distribution $P(\y)$, by $\sigma := \sigma(t)$ the time-dependent noise level, and by $\x := \x(t)$ a point along a denoising trajectory. We set $\sigma(0)=0$ and choose with $\sigma(1)=80$ a value such that $\sigma(t)\gg std(y_i)$, for all dimensions $i\in\{1,2,..,d\}$. We assume $\sigma(t)$ to be a strictly monotone increasing function in time $t$. To derive the optimal denoiser, we make use of the ordinary differential equation (ODE) that describes the field lines of probability flow (e.g. Eq. (1) of Ref. \cite{karras_edm}). This ODE allows for the generation of sampling trajectories 
\begin{equation} \label{eq:ode}
    d\x = -\dot{\sigma} \sigma \nabla_{\x} \log P(\x| \sigma^2) dt ,
\end{equation}
when initialized with a noise sample $\x(1)\sim {\mathcal N}(\x|0,\sigma(1)^2)$. Here and in following we use the definition $\mathcal{N}(\x|\bm \mu, \sigma^2) := \mathcal{N}(\x|\bm \mu, \sigma^2 I_d)$ to denote an isotropic normal distribution. Note that $dt<0$ for inference, by definition.
Futhermore, $\dot{\sigma} \sigma$ can be interpreted as a diffusion coefficient, $\nabla_{\x} \log P(\x| \sigma^2)$ as a score function, %
and $P(\x| \sigma^2) = \int_{\mathbb{R}^d} \mathcal{N}(\x|\y, \sigma^2) P(\y) d\y$ as the distribution of noisy samples, which results from adding Gaussian noise to data samples. To derive the optimal denoiser, we rewrite the score function as
\begin{align} \label{eq:score-transform}
    -\nabla_{\x} \log P(\x| \sigma^2) 
    &= -\frac{\nabla_{\x} P(\x| \sigma^2)}{P(\x| \sigma^2)} \nonumber \\
    &= -\frac{\nabla_{\x} \int_{\mathbb{R}^d} \mathcal{N}(\x|\y, \sigma^2) P(\y) d\y}{P(\x| \sigma^2)} \nonumber \\
    &= \frac{\int_{\mathbb{R}^d} \frac{\x-\y}{\sigma^2}\mathcal{N}(\x|\y, \sigma^2) P(\y) d\y}{P(\x| \sigma^2)} \nonumber \\
    &= \frac{1}{\sigma^2} \int_{\mathbb{R}^d} (\x-\y) P(\y|\x, \sigma^2)d\y \nonumber \\
    &= \frac{1}{\sigma^2} \bigg( \int_{\mathbb{R}^d} \x P(\y|\x, \sigma^2)d\y - \int_{\mathbb{R}^d} \y P(\y|\x, \sigma^2)d\y \bigg) \nonumber \\
    &= \frac{\x - \y^*(\x, \sigma)}{\sigma^2} ,
\end{align}
where the optimal denoiser is given by the posterior mean
\begin{equation} \label{eq:optimal-denoiser}
    \y^*(\x, \sigma) 
    := \int_{\mathbb{R}^d} \y P(\y|\x, \sigma^2)d\y 
    = \int_{\mathbb{R}^d} \y \frac{\mathcal{N}(\x|\y, \sigma^2)P(\y)}{P(\x | \sigma^2)} d\y . 
\end{equation}
To construct an error-prone denoiser, we substitute the true data distribution with a ``broader'' data distribution $P(\y)\rightarrow P_\delta(\y)$, with $P_\delta(\y) = \int_{\mathbb{R}^d} \mathcal{N}(\y|\y',\delta^2) P(\y') d\y'$, where $\delta$ is the standard deviation of noise that randomly shifts the true data points. The corresponding posterior mean is given by
\begin{align} \label{errdenoiser}
    \y_\delta(\x, \sigma) 
    :=& \int_{\mathbb{R}^d} \y \frac{\mathcal{N}(\x|\y, \sigma^2) P_\delta(\y)}{P_\delta(\x | \sigma^2)} d\y \nonumber \\
    =&\frac{\int_{\mathbb{R}^d}\int_{\mathbb{R}^d} \y \,\mathcal{N}(\x|\y, \sigma^2)  \mathcal{N}(\y|\y',\delta^2) P(\y') d\y' d\y}{\int_{\mathbb{R}^d}\int_{\mathbb{R}^d}  \mathcal{N}(\x|\y, \sigma^2)  \mathcal{N}(\y|\y',\delta^2) P(\y') d\y' d\y} \quad,
\end{align}
with  $P_{\delta}(\x| \sigma^2) := \int_{\mathbb{R}^d} \mathcal{N}(\x|\y, \sigma^2) P_{\delta}(\y) d\y$.
To find an explicit expression for $\y_\delta$, we utilize known integrals \cite{owen1980integrals}.
For the denominator, we use the expression 
\begin{equation} \label{eq:formula1}
    \int_{\mathbb{R}^d} \phi(\x) \phi({\bm a} + b\x) d\x 
    = \frac{1}{\sqrt{1+b^2}} \phi \bigg(\frac{{\bm a}}{\sqrt{1+b^2}} \bigg) .
\end{equation}
with $\phi(\x):={\cal N}(\x|0,1)$. However, we have to reformulate this expression for the product of two Gaussians with different means and variances. To this end, consider the substitution $\x = \frac{\z-{\bm c}}{d}, c, d \in \mathbb{R}$ and adjust the LHS.
\begin{align*}
    \int_{\mathbb{R}^d} \phi(\x) \phi({\bm a} + b\x) d\x
    = &\int_{\mathbb{R}^d} \phi\bigg(\frac{\z-{\bm c}}{d}\bigg) \phi\bigg({\bm a} + b\frac{\z-{\bm c}}{d}\bigg) \frac{1}{d} d\z \\
    = &\int_{\mathbb{R}^d} \phi\bigg(\frac{\z-{\bm c}}{d}\bigg) \phi\bigg(\frac{\z - {\bm \mu}}{\sigma}\bigg) \frac{1}{d} d\z \qquad , \ {\bm \mu} = {\bm c} - {\bm a}\frac{d}{b}, \sigma = \frac{d}{b} \\
    = &\sigma \int_{\mathbb{R}^d} \mathcal{N}(\z | {\bm c}, d^2) \mathcal{N}(\z | {\bm \mu}, \sigma^2) d\z .
\end{align*}
Combining with the RHS of \Cref{eq:formula1} yields the desired integration formula
\begin{align} \label{eq:denominator}
    \int_{\mathbb{R}^d} \mathcal{N}(\z | {\bm c}, d^2) \mathcal{N}(\z | {\bm \mu}, \sigma^2) d\z 
    = &\frac{1}{\sigma} \frac{1}{\sqrt{1+b^2}} \phi \bigg(\frac{{\bm a}}{\sqrt{1+b^2}} \bigg) \nonumber \\
    = &\frac{1}{\sqrt{\sigma^2 + d^2}} \phi \bigg(\frac{{\bm c} -{\bm \mu}}{\sqrt{\sigma^2 + d^2}} \bigg) 
    = \mathcal{N}({\bm c}| {\bm \mu}, \sigma^2+d^2) .
\end{align}
For the numerator, we use the expression \cite{owen1980integrals}
\begin{equation} \label{eq:formula2}
    \int_{\mathbb{R}^d} \x \phi(\x) \phi({\bm a} + b\x) d\x
    = -\frac{{\bm a}b}{(1+b^2)^\frac{3}{2}} \phi \bigg(\frac{{\bm a}}{\sqrt{1+b^2}} \bigg) .
\end{equation}
Using the same substitutions as before yields
\begin{align*}
    &\int_{\mathbb{R}^d} \x \phi(\x) \phi({\bm a} + b\x) d\x  \\
    = &\int_{\mathbb{R}^d} \frac{\z -{\bm c}}{d} \phi\bigg(\frac{\z-{\bm c}}{d}\bigg) \phi\bigg(\frac{\z - {\bm \mu}}{\sigma}\bigg) \frac{1}{d} d\z \\
    = &\frac{1}{d} \int_{\mathbb{R}^d} \z \sigma \mathcal{N}(\z | {\bm c}, d^2) \mathcal{N}(\z | {\bm \mu}, \sigma^2) d\z- \frac{c}{d} \int_{\mathbb{R}^d} \mathcal{N}(\z | {\bm c}, d^2) \mathcal{N}(\z | {\bm \mu}, \sigma^2) d\z \\
    \overset{(\ref{eq:denominator})}{=} &\frac{1}{d} \int_{\mathbb{R}^d} \z \sigma \mathcal{N}(\z | {\bm c}, d^2) \mathcal{N}(\z | {\bm \mu}, \sigma^2) d\z - \frac{c}{d} \mathcal{N}({\bm c}| {\bm \mu}, \sigma^2 + d^2) .
\end{align*}
Combining with the RHS of \Cref{eq:formula2} yields the second integration formula
\begin{align} \label{eq:numerator}
    &\int_{\mathbb{R}^d} \z \sigma \mathcal{N}(\z | {\bm c}, d^2) \mathcal{N}(\z | {\bm \mu}, \sigma^2) d\z \\
    = &\frac{d}{\sigma} \Bigg( - \frac{{\bm c} -{\bm \mu}}{\sigma} \frac{d}{\sigma} \bigg( \frac{\sigma}{\sqrt{\sigma^2 + d^2}} \bigg)^3 \phi \bigg( \frac{{\bm c} -{\bm \mu}}{\sqrt{\sigma^2 + d^2}} \bigg)
    + \frac{\sigma {\bm c}}{d} \mathcal{N}({\bm c}|{\bm \mu}, \sigma^2 + d^2) \Bigg) \nonumber \\
    = &\frac{d}{\sigma} \Bigg( - \frac{({\bm c} -{\bm \mu})d}{\sigma^2} \frac{\sigma^3}{\sigma^2 + d^2} \mathcal{N}({\bm c}|{\bm \mu}, \sigma^2 + d^2)
    + \frac{\sigma {\bm c}}{d} \mathcal{N}({\bm c}|{\bm \mu}, \sigma^2 + d^2) \Bigg) \nonumber \\
    = &\mathcal{N}({\bm c}|{\bm \mu}, \sigma^2 + d^2) \bigg( {\bm c} - \frac{({\bm c} -{\bm \mu}) d^2}{\sigma^2 + d^2} \bigg) \nonumber \\
    = &\mathcal{N}({\bm c}|{\bm \mu}, \sigma^2 + d^2) \frac{{\bm c} \sigma^2 + {\bm \mu} d^2}{\sigma^2 + d^2} .
\end{align}
For the numerator, of Eq.~(\ref{errdenoiser}) we now get 
\begin{align} \label{eq:sol-numerator}
    &\int_{\mathbb{R}^d} \y \mathcal{N}(\x|\y, \sigma^2) \int_{\mathbb{R}^d} \mathcal{N}(\y|\y',\delta^2) P(\y') d\y' d\y \\
    = & \int_{\mathbb{R}^d} \int_{\mathbb{R}^d} \y \mathcal{N}(\y|\x, \sigma^2) \mathcal{N}(\y|\y',\delta^2) d\y P(\y') d\y' \nonumber \\
    \overset{(\ref{eq:numerator})}{=} & \int_{\mathbb{R}^d} \mathcal{N}(\x|\y', \sigma^2 + \delta^2) \frac{\x\delta^2 + \y' \sigma^2}{\sigma^2 + \delta^2} P(\y') d\y' \nonumber \\
    = & \frac{1}{\sigma^2 + \delta^2} \int_{\mathbb{R}^d} (\x\delta^2 + \y' \sigma^2) P(\x,\y'| \sigma^2 + \delta^2) d\y'  .
\end{align}
For the denominator of Eq.~(\ref{errdenoiser}), we get 
\begin{align} \label{eq:sol-denominator}
    P_\delta(\x | \sigma^2)
    = & \int_{\mathbb{R}^d} \mathcal{N}(\x|\y, \sigma^2) P_\delta(\y) d\y \nonumber \\
    = & \int_{\mathbb{R}^d} \mathcal{N}(\x|\y, \sigma^2) \int_{\mathbb{R}^d} \mathcal{N}(\y|\y',\delta^2) P(\y') d\y' d\y \nonumber \\
    = & \int_{\mathbb{R}^d} \int_{\mathbb{R}^d} \mathcal{N}(\y|\x, \sigma^2) \mathcal{N}(\y|\y',\delta^2) d\y P(\y') d\y' \nonumber \\
    \overset{(\ref{eq:denominator})}{=} & \int_{\mathbb{R}^d} \mathcal{N}(\x | \y', \sigma^2+\delta^2) P(\y') d\y' \nonumber \\
    = & \int_{\mathbb{R}^d} \mathcal{N}(\x | \y', \sigma^2+\delta^2) P(\y') d\y' \nonumber \\
    = & P(\x|\sigma^2+\delta^2) .
\end{align}
Finally,
\begin{align} \label{eq:sol-error-denoiser}
    \y_\delta(\x, \sigma) 
    \overset{(\ref{eq:sol-numerator}) + (\ref{eq:sol-denominator})}{=} & \frac{\int_{\mathbb{R}^d} (\x\delta^2 + \y' \sigma^2) P(\x,\y'| {\tilde \sigma}^2) d\y'}{P(\x|{\tilde \sigma}^2)} \frac{1}{{\tilde \sigma}^2} \nonumber \\
    = & \frac{\x\delta^2 \int_{\mathbb{R}^d} P(\x,\y'| {\tilde \sigma}^2) d\y' + \sigma^2 \int_{\mathbb{R}^d} \y' P(\x,\y'| {\tilde \sigma}^2) d\y'}{P(\x|{\tilde \sigma}^2)} \frac{1}{{\tilde \sigma}^2} \nonumber \\
    = & \frac{\x\delta^2  + \sigma^2 \int_{\mathbb{R}^d} \y' \frac{P(\x,\y'| {\tilde \sigma}^2)}{P(\x|{\tilde \sigma}^2)} d\y'}{{\tilde \sigma}^2}  \nonumber \\
    = &\frac{\x\delta^2 + \y^*(\x, {\tilde\sigma}) \sigma^2}{{\tilde \sigma}^2} .
\end{align}
with ${\tilde \sigma}^2 = \sigma^2 + \delta^2$. Using the optimal denoiser $\y^*$, we define the optimal noise predictor
\begin{equation} \label{eq:optimal-noise-predictor}
    \e^*(\x,\sigma)
    := -\sigma \nabla_{\x} \log P(\x| \sigma^2)
    \overset{(\ref{eq:score-transform})}{=} \frac{\x - \y^*(\x, \sigma)}{\sigma} ,
\end{equation}
and similarly, we define the error-prone noise predictor as
\begin{equation} \label{eq:error-noise-predictor}
    \e_{err}(\x,\sigma)
    := \frac{\x - \y_\delta(\x, \sigma)}{\sigma} .
\end{equation}
Inserting the explict expression for $\y_\delta$ in \Cref{eq:error-noise-predictor} yields the expression for $\e_{err}$ used in the main text
\[
    \e_{err}(\x, \sigma) 
    = \frac{\x - \y_\delta(\x, \sigma)}{\sigma}
    = \frac{\x - \frac{\x\delta^2 + \y^*(\x, {\tilde\sigma}) \sigma^2}{{\tilde \sigma}^2}}{\sigma}
    = \frac{1}{\sigma} \frac{\sigma^2 (\x - \y^*(\x, {\tilde\sigma}))}{{\tilde \sigma}^2}
    = \frac{\sigma}{{\tilde \sigma}} \e^*(\x, {\tilde\sigma}) .
\]
We can now generate trajectories $\x(t)$ from \Cref{eq:ode} by numerically solving the ODE with $\y^*$ or $\y_\delta$, corresponding to $\e^*$ or $\e_{err}$. %
In practice, we consider a finite set of data-points $\{\y_i\}_{i=1}^N$, in which case \Cref{eq:optimal-denoiser} and \Cref{eq:sol-error-denoiser} simplify to
\begin{align}
    \y^*(\x, \sigma) &= \frac{\sum_i \y_i N(\x(t)|\y_i,\sigma^2)}{\sum_i N(\x(t)|\y_i,\sigma^2)} \\
    \y_\delta(\x, \sigma)  &= \frac{\x\delta^2 + \y^*(\x, {\tilde\sigma}) \sigma^2}{{\tilde \sigma}^2} ,
\end{align}
which can be directly used to simulate trajectories. To do so, we use the Euler method shown in \Cref{alg:euler}. We found that higher-order integration methods (e.g., Alg.~1 of Ref.  \cite{karras_edm}) can be used but are not necessary to achieve sufficient accuracy. This framework can be used for simulations of different configurations of $\e^*$, $\e_{err}$, conditional and unconditional models, as well as arbitrary finite datasets in $\mathbb{R}^n$. \Cref{fig:toy_grid_triangle,fig:toy_grid_cloud} show a detailed analysis of WMG on two datasets, the triangle formation used in the main text, and a cloud of normally distributed data samples.

\begin{algorithm}
\caption{Euler method} \label{alg:euler}
\begin{algorithmic}
    \Require positive denoiser $\y_\delta$, negative denoiser $\y_{\delta'}$, guidance weight $w(\x, \sigma)$, $t_{i \in \{0,...,N\}} \in [0,1]$
    \State \textbf{sample} $x_0 \sim \mathcal{N}(\x | 0, \sigma(1)^2 I_d)$
    \For{$i$ in $0,...,N-1$}
        \State $\sigma_i, \sigma_{i+1} = \sigma(t_i), \sigma(t_{i+1})$  \Comment{Set current and next noise level}
        \State $\tilde \y_i = \y_\delta(\x_i, \sigma_i^2) + w(\x, \sigma_i) [\y_\delta(\x_i, \sigma_i^2) - \y_{\delta'}(\x_i, \sigma_i^2)]$ \Comment{Compute guidance target prediction}
        \State $\bm d_i = (\x_i - \tilde \y_i) / \sigma_i$  \Comment{Compute score}
        \State $\x_{i+1} = \x_i + (\sigma_{i+1} - \sigma_i) \bm d_i$  \Comment{Update the trajectory}
    \EndFor
\end{algorithmic}
\end{algorithm}

\vfill
\input{figs/toy_opt_weight}
\vspace*{\fill}

\newpage
\input{figs/toy_full_grid1}
\input{figs/toy_full_grid_cloud}
\clearpage

%% file: figs/toy_opt_weight.tex
\begin{figure*}[h]
    \centering
    \begin{subfigure}[b]{0.48\textwidth}
        \includegraphics[width=1\textwidth]{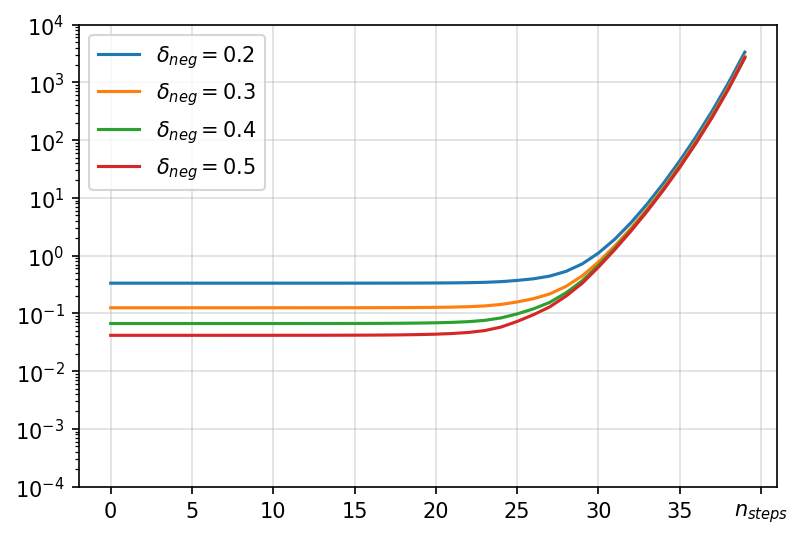}
        \caption{\wmg{}, $\delta_\text{pos} = 0.1$}
    \end{subfigure}
    \begin{subfigure}[b]{0.48\textwidth}
        \includegraphics[width=1\textwidth]{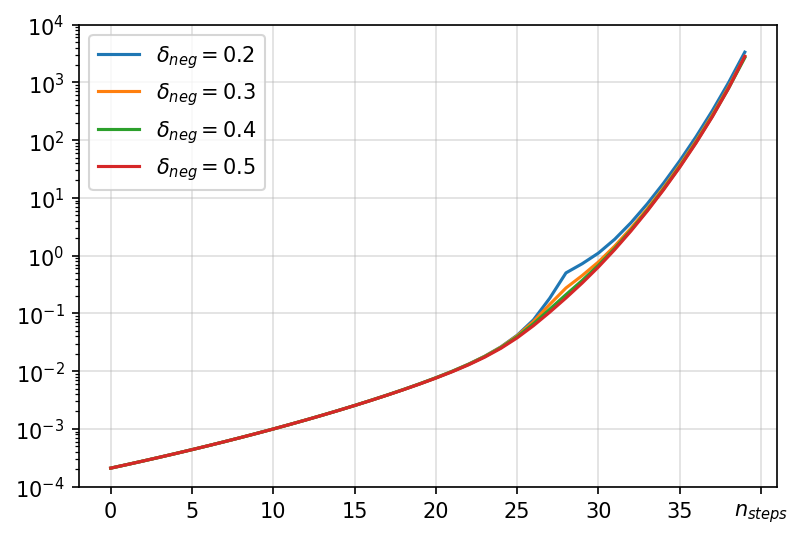}
        \caption{CFG, $\delta_\text{pos} = 0.1$}
    \end{subfigure}
    \caption{Optimal guidance weight for WMG and CFG in a toy example simulation. The $x$-axis shows the sampling step and the $y$-axis the guidance weight value, averaged over 100 trajectories with random initialization. The dataset and models $\e_{\text{pos}}$ and $\e_{\text{neg}}$ are as described in the main text. For both guidance methods, $\e_{\text{pos}}$ has a variance of $\delta_{\text{pos}} = 0.1$ and different values for $\delta_{\text{neg}}$ are shown.}
\end{figure*}

%% file: figs/toy_full_grid1.tex
\begin{figure*}
    \centering
    \includegraphics[width=1\textwidth]{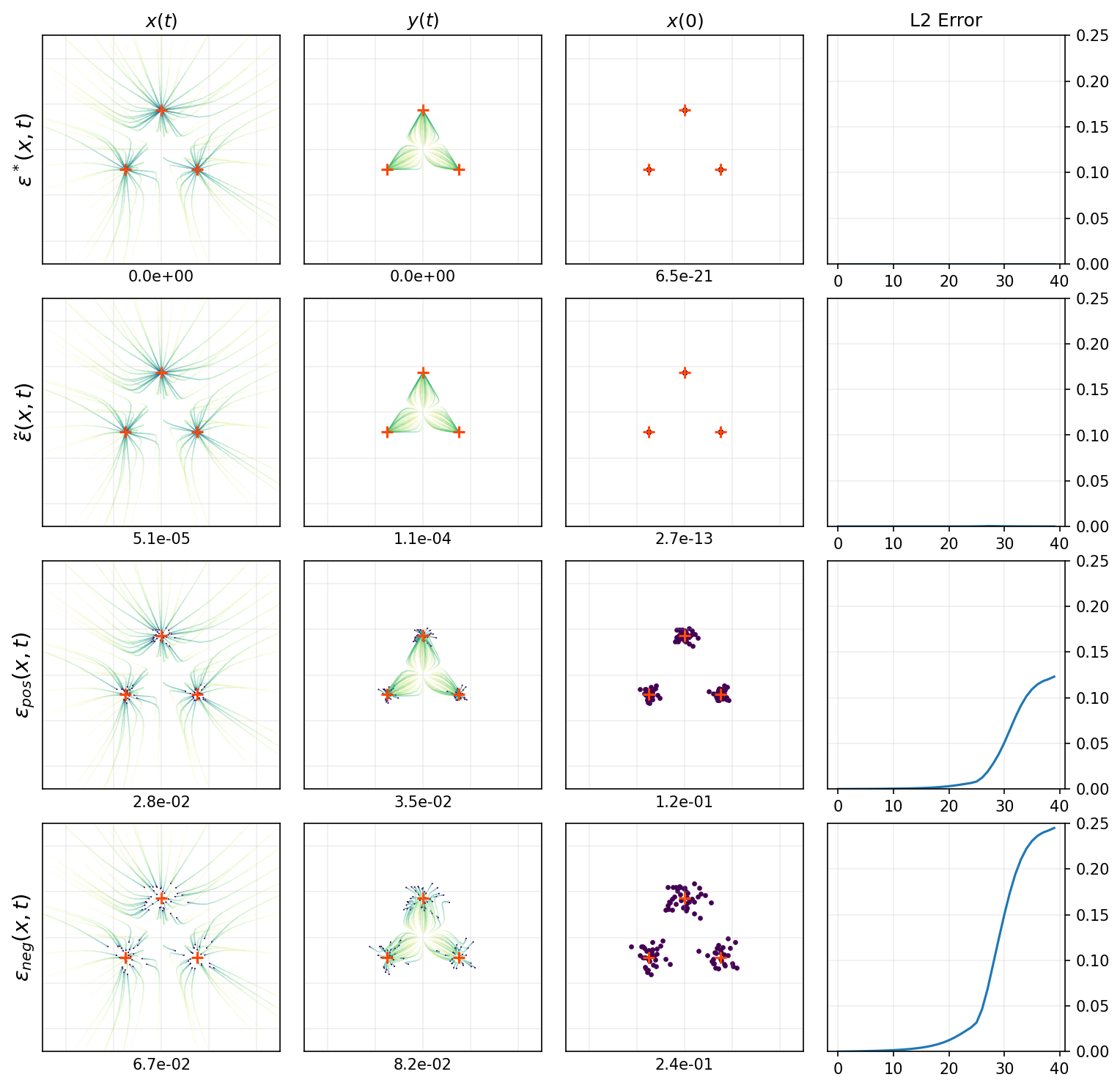}
    \caption{Toy example simulation for WMG with optimal guidance weight $w^*(\x,t)$ for three equidistant data points, using 40 sampling steps. \textbf{Columns from left to right:} The trajectory $\x(t)$, the target prediction $\y(t)$, the trajectory endpoints $\x(1)$, and the average step-wise L2 error to the optimal noise predictor $\e^*$ along the trajectory. The numbers below the plots in the first three columns show the average L2 error to the optimal trajectory (additionally averaged over steps for $\x(t)$ and $\y(t)$). \textbf{Rows from top to bottom:} The optimal noise predictor $\e^*$, the WMG noise predictor $\tilde \e = \e_{\text{pos}} + w [\e_{\text{pos}} - \e_{\text{neg}}]$ with optimal weight $w^*(\x, t)$, $\e_{\text{pos}}$ ($w=0$), and $\e_{\text{neg}}$ ($w=-1$).}
    \label{fig:toy_grid_triangle}
\end{figure*}

%% file: figs/toy_full_grid_cloud.tex
\begin{figure*}
    \centering
    \includegraphics[width=1\textwidth]{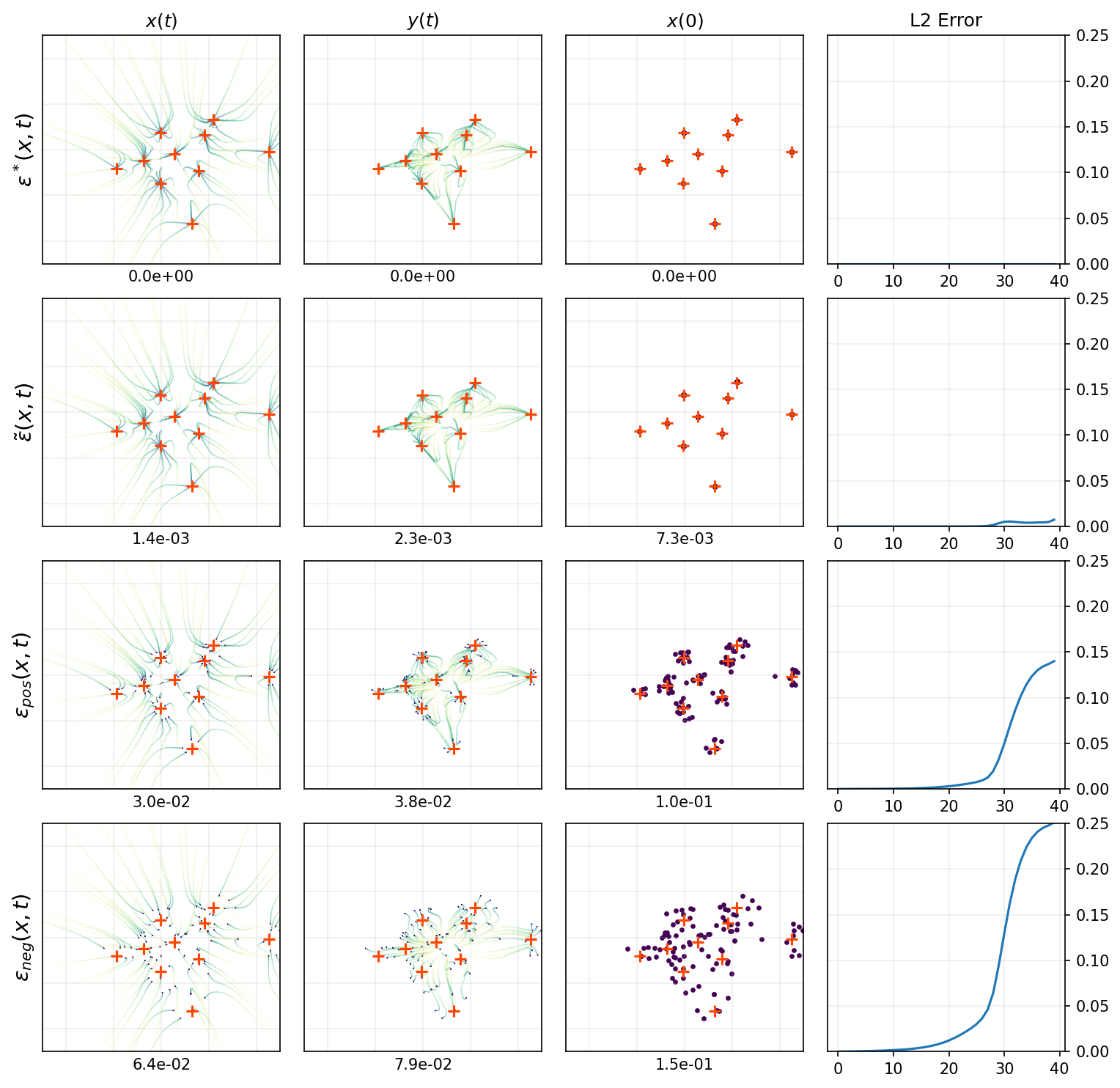}
    \caption{Toy example simulations for WMG with optimal guidance weight $w^*(\x,t)$ for cloud of normally distributed data points, using 40 sampling steps. \textbf{Columns from left to right:} The trajectory $\x(t)$, the target prediction $\y(t)$, the trajectory endpoints $\x(1)$, and the average step-wise L2 error to the optimal noise predictor $\e^*$ along the trajectory. The numbers below the plots in the first three columns show the average L2 error to the optimal trajectory (additionally averaged over steps for $\x(t)$ and $\y(t)$). \textbf{Rows from top to bottom:} The optimal noise predictor $\e^*$, the WMG noise predictor $\tilde \e = \e_{\text{pos}} + w [\e_{\text{pos}} - \e_{\text{neg}}]$ with optimal weight $w^*(\x, t)$, $\e_{\text{pos}}$ ($w=0$), and $\e_{\text{neg}}$ ($w=-1$).}
    \label{fig:toy_grid_cloud}
\end{figure*}

%% file: fig_wrap/rct_swg_supp.tex
\begin{figure*}
\centering\includegraphics[width=0.7\linewidth]{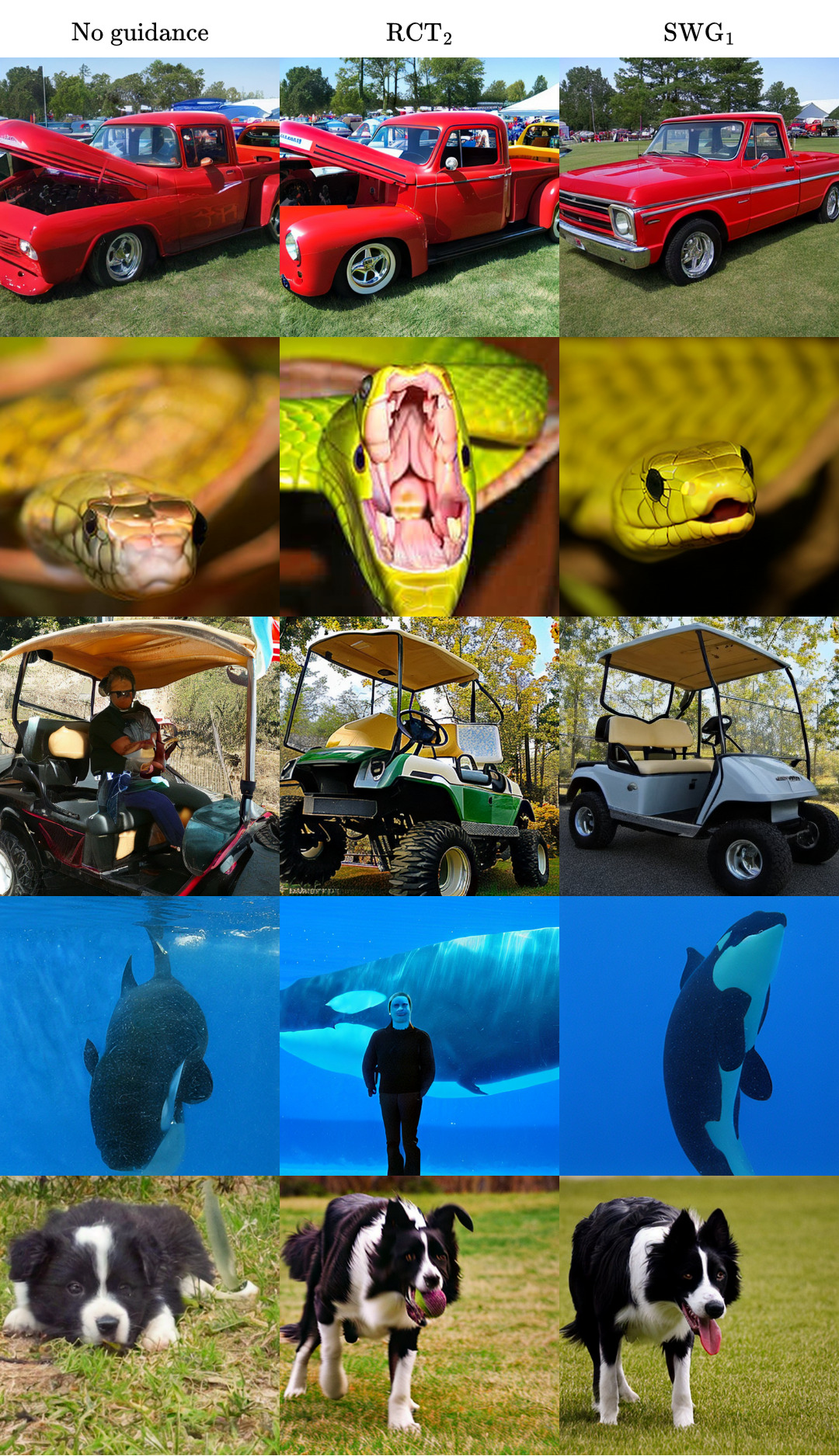}
    \caption{Generated samples from the human evaluation where RCT introduced artifacts.}    \label{fig:rct_vs_swg_triplets_001}
\end{figure*}

%% file: fig_wrap/rct_swg_guidance_scales.tex
\begin{figure*}
\centering\includegraphics[width=0.99\linewidth]{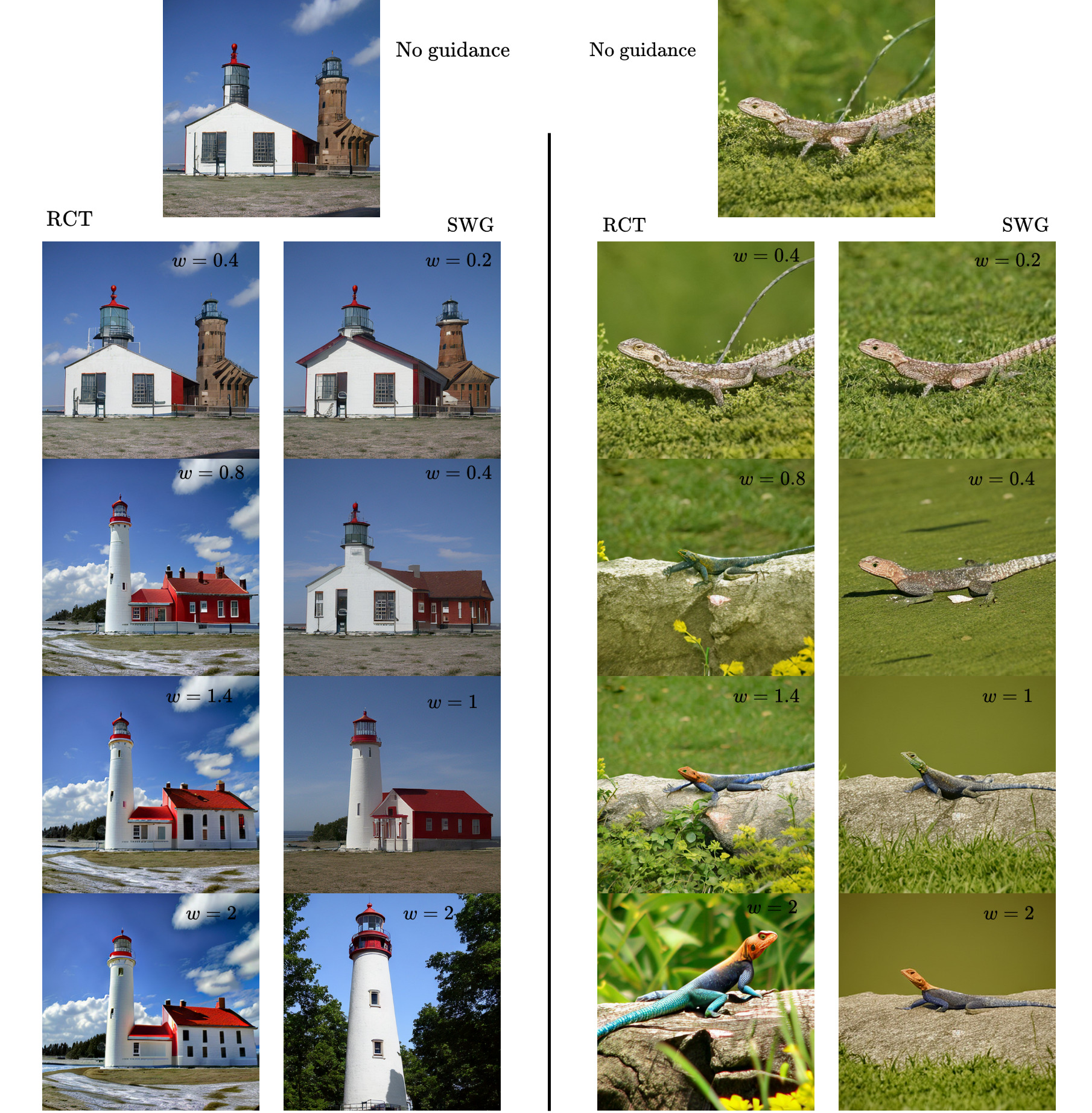}
    \caption{More examples with different guidance scales $w$ for RCT (left) and SWG (right).} \label{fig:rct_vs_swg_supp_guidance_scales_w_002}   
\end{figure*}

%% file: fig_wrap/teaser_supp.tex
\begin{figure*}
\centering\includegraphics[width=0.7\linewidth]{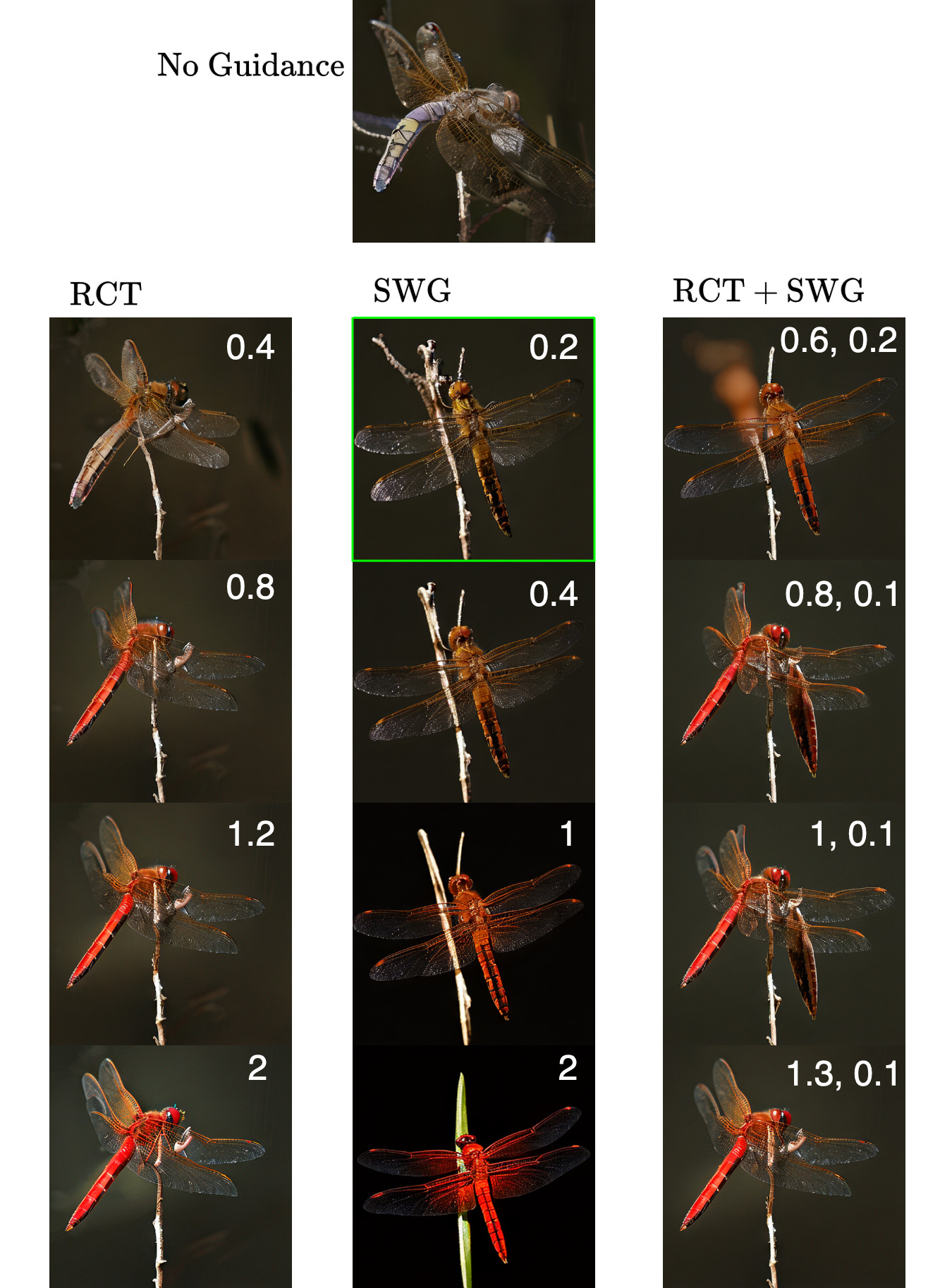}
    \caption{Generated EDM2-XXL 512$\times$512 samples using no guidance, RCT (left), SWG (center), and RCT+SWG (right). The guidance scales are shown in white on the top right corner of the image. The green box shows the sample that was used in the main paper, Figure 1.}    \label{fig:failures_teser_006}
\end{figure*}

%% file: main.bbl
\begin{thebibliography}{50}
\providecommand{\natexlab}[1]{#1}
\providecommand{\url}[1]{\texttt{#1}}
\expandafter\ifx\csname urlstyle\endcsname\relax
  \providecommand{\doi}[1]{doi: #1}\else
  \providecommand{\doi}{doi: \begingroup \urlstyle{rm}\Url}\fi

\bibitem[Adaloglou et~al.(2025)Adaloglou, Kaiser, Michels, and Kollmann]{adaloglou2025rethinking}
Nikolas Adaloglou, Tim Kaiser, Felix Michels, and Markus Kollmann.
\newblock Rethinking cluster-conditioned diffusion models for label-free image synthesis.
\newblock In \emph{2025 IEEE/CVF Winter Conference on Applications of Computer Vision (WACV)}, pages 3603--3613. IEEE, 2025.

\bibitem[Ahn et~al.(2024)Ahn, Cho, Min, Jang, Kim, Kim, Park, Jin, and Kim]{ahn2024pag}
Donghoon Ahn, Hyoungwon Cho, Jaewon Min, Wooseok Jang, Jungwoo Kim, SeonHwa Kim, Hyun~Hee Park, Kyong~Hwan Jin, and Seungryong Kim.
\newblock Self-rectifying diffusion sampling with perturbed-attention guidance.
\newblock \emph{arXiv preprint arXiv:2403.17377}, 2024.

\bibitem[Ahsan et~al.(2024)Ahsan, Raman, Liu, and Siddique]{ahsan2024comprehensive}
Md~Manjurul Ahsan, Shivakumar Raman, Yingtao Liu, and Zahed Siddique.
\newblock A comprehensive survey on diffusion models and their applications.
\newblock 2024.

\bibitem[Alemohammad et~al.(2024)Alemohammad, Humayun, Agarwal, Collomosse, and Baraniuk]{alemohammad2024sims}
Sina Alemohammad, Ahmed~Imtiaz Humayun, Shruti Agarwal, John Collomosse, and Richard Baraniuk.
\newblock Self-improving diffusion models with synthetic data.
\newblock \emph{arXiv preprint arXiv:2408.16333}, 2024.

\bibitem[Balaji et~al.(2022)Balaji, Nah, Huang, Vahdat, Song, Zhang, Kreis, Aittala, Aila, Laine, et~al.]{balaji2022ediff_att}
Yogesh Balaji, Seungjun Nah, Xun Huang, Arash Vahdat, Jiaming Song, Qinsheng Zhang, Karsten Kreis, Miika Aittala, Timo Aila, Samuli Laine, et~al.
\newblock ediff-i: Text-to-image diffusion models with an ensemble of expert denoisers.
\newblock \emph{arXiv preprint arXiv:2211.01324}, 2022.

\bibitem[Bar-Tal et~al.(2023)Bar-Tal, Yariv, Lipman, and Dekel]{bar2023multidiffusion}
Omer Bar-Tal, Lior Yariv, Yaron Lipman, and Tali Dekel.
\newblock Multidiffusion: Fusing diffusion paths for controlled image generation.
\newblock 2023.

\bibitem[Bradley and Nakkiran(2024)]{bradley2024classifier}
Arwen Bradley and Preetum Nakkiran.
\newblock Classifier-free guidance is a predictor-corrector.
\newblock \emph{arXiv preprint arXiv:2408.09000}, 2024.

\bibitem[Cardoso et~al.(2022)Cardoso, Li, Brown, Ma, Kerfoot, Wang, Murrey, Myronenko, Zhao, Yang, et~al.]{cardoso2022monai}
M~Jorge Cardoso, Wenqi Li, Richard Brown, Nic Ma, Eric Kerfoot, Yiheng Wang, Benjamin Murrey, Andriy Myronenko, Can Zhao, Dong Yang, et~al.
\newblock Monai: An open-source framework for deep learning in healthcare.
\newblock \emph{arXiv preprint arXiv:2211.02701}, 2022.

\bibitem[Chong and Forsyth(2020)]{chong2020effectivelyunbiasedfidinception}
Min~Jin Chong and David Forsyth.
\newblock Effectively unbiased fid and inception score and where to find them, 2020.
\newblock URL \url{https://arxiv.org/abs/1911.07023}.

\bibitem[Chung et~al.(2024)Chung, Kim, Park, Nam, and Ye]{chung2024cfg++}
Hyungjin Chung, Jeongsol Kim, Geon~Yeong Park, Hyelin Nam, and Jong~Chul Ye.
\newblock Cfg++: Manifold-constrained classifier free guidance for diffusion models.
\newblock \emph{arXiv preprint arXiv:2406.08070}, 2024.

\bibitem[Dhariwal and Nichol(2021)]{dhariwal2021diffusion}
Prafulla Dhariwal and Alexander Nichol.
\newblock Diffusion models beat gans on image synthesis.
\newblock \emph{Advances in neural information processing systems}, 34:\penalty0 8780--8794, 2021.

\bibitem[Dieleman(2025)]{dieleman2025latents}
Sander Dieleman.
\newblock Generative modelling in latent space, 2025.
\newblock URL \url{https://sander.ai/2025/04/15/latents.html}.

\bibitem[Hertz et~al.(2022)Hertz, Mokady, Tenenbaum, Aberman, Pritch, and Cohen-Or]{hertz2022prompt_att}
Amir Hertz, Ron Mokady, Jay Tenenbaum, Kfir Aberman, Yael Pritch, and Daniel Cohen-Or.
\newblock Prompt-to-prompt image editing with cross attention control.
\newblock \emph{arXiv preprint arXiv:2208.01626}, 2022.

\bibitem[Heusel et~al.(2018)Heusel, Ramsauer, Unterthiner, Nessler, and Hochreiter]{fid_orig}
Martin Heusel, Hubert Ramsauer, Thomas Unterthiner, Bernhard Nessler, and Sepp Hochreiter.
\newblock Gans trained by a two time-scale update rule converge to a local nash equilibrium, 2018.

\bibitem[Ho and Salimans(2022)]{ho2022classifierfreediffusionguidance}
Jonathan Ho and Tim Salimans.
\newblock Classifier-free diffusion guidance, 2022.
\newblock URL \url{https://arxiv.org/abs/2207.12598}.

\bibitem[Ho et~al.(2020)Ho, Jain, and Abbeel]{ho2020denoising}
Jonathan Ho, Ajay Jain, and Pieter Abbeel.
\newblock Denoising diffusion probabilistic models, 2020.

\bibitem[Hong(2024)]{hong2024smoothed}
Susung Hong.
\newblock Smoothed energy guidance: Guiding diffusion models with reduced energy curvature of attention.
\newblock \emph{arXiv preprint arXiv:2408.00760}, 2024.

\bibitem[Hong et~al.(2023)Hong, Lee, Jang, and Kim]{hong2023sag}
Susung Hong, Gyuseong Lee, Wooseok Jang, and Seungryong Kim.
\newblock Improving sample quality of diffusion models using self-attention guidance.
\newblock In \emph{Proceedings of the IEEE/CVF International Conference on Computer Vision}, pages 7462--7471, 2023.

\bibitem[Jayasumana et~al.(2024)Jayasumana, Ramalingam, Veit, Glasner, Chakrabarti, and Kumar]{jayasumana2024rethinking_fid}
Sadeep Jayasumana, Srikumar Ramalingam, Andreas Veit, Daniel Glasner, Ayan Chakrabarti, and Sanjiv Kumar.
\newblock Rethinking fid: Towards a better evaluation metric for image generation.
\newblock In \emph{Proceedings of the IEEE/CVF Conference on Computer Vision and Pattern Recognition}, pages 9307--9315, 2024.

\bibitem[Jiang et~al.(2023)Jiang, Sablayrolles, Mensch, Bamford, Chaplot, de~las Casas, Bressand, Lengyel, Lample, Saulnier, et~al.]{jiang2023mistral}
AQ~Jiang, A~Sablayrolles, A~Mensch, C~Bamford, DS~Chaplot, D~de~las Casas, F~Bressand, G~Lengyel, G~Lample, L~Saulnier, et~al.
\newblock Mistral 7b (2023).
\newblock \emph{arXiv preprint arXiv:2310.06825}, 2023.

\bibitem[Kaplan et~al.(2020)Kaplan, McCandlish, Henighan, Brown, Chess, Child, Gray, Radford, Wu, and Amodei]{kaplan2020scaling}
Jared Kaplan, Sam McCandlish, Tom Henighan, Tom~B Brown, Benjamin Chess, Rewon Child, Scott Gray, Alec Radford, Jeffrey Wu, and Dario Amodei.
\newblock Scaling laws for neural language models.
\newblock \emph{arXiv preprint arXiv:2001.08361}, 2020.

\bibitem[Karras et~al.(2022)Karras, Aittala, Aila, and Laine]{karras_edm}
Tero Karras, Miika Aittala, Timo Aila, and Samuli Laine.
\newblock Elucidating the design space of diffusion-based generative models, 2022.

\bibitem[Karras et~al.(2024{\natexlab{a}})Karras, Aittala, Kynkäänniemi, Lehtinen, Aila, and Laine]{karras2024guidingdiffusionmodelbad}
Tero Karras, Miika Aittala, Tuomas Kynkäänniemi, Jaakko Lehtinen, Timo Aila, and Samuli Laine.
\newblock Guiding a diffusion model with a bad version of itself, 2024{\natexlab{a}}.
\newblock URL \url{https://arxiv.org/abs/2406.02507}.

\bibitem[Karras et~al.(2024{\natexlab{b}})Karras, Aittala, Lehtinen, Hellsten, Aila, and Laine]{karras2024analyzingimprovingtrainingdynamics}
Tero Karras, Miika Aittala, Jaakko Lehtinen, Janne Hellsten, Timo Aila, and Samuli Laine.
\newblock Analyzing and improving the training dynamics of diffusion models, 2024{\natexlab{b}}.
\newblock URL \url{https://arxiv.org/abs/2312.02696}.

\bibitem[Kingma et~al.(2023)Kingma, Salimans, Poole, and Ho]{kingma2023variational}
Diederik~P. Kingma, Tim Salimans, Ben Poole, and Jonathan Ho.
\newblock Variational diffusion models, 2023.

\bibitem[Kouzelis et~al.(2025)Kouzelis, Kakogeorgiou, Gidaris, and Komodakis]{kouzelis2025eq}
Theodoros Kouzelis, Ioannis Kakogeorgiou, Spyros Gidaris, and Nikos Komodakis.
\newblock Eq-vae: Equivariance regularized latent space for improved generative image modeling.
\newblock \emph{arXiv preprint arXiv:2502.09509}, 2025.

\bibitem[Kynkäänniemi et~al.(2023)Kynkäänniemi, Karras, Aittala, Aila, and Lehtinen]{kynkäänniemi2023role}
Tuomas Kynkäänniemi, Tero Karras, Miika Aittala, Timo Aila, and Jaakko Lehtinen.
\newblock The role of imagenet classes in fr\'echet inception distance, 2023.

\bibitem[Kynkäänniemi et~al.(2024)Kynkäänniemi, Aittala, Karras, Laine, Aila, and Lehtinen]{kynkäänniemi2024applyingguidancelimitedinterval}
Tuomas Kynkäänniemi, Miika Aittala, Tero Karras, Samuli Laine, Timo Aila, and Jaakko Lehtinen.
\newblock Applying guidance in a limited interval improves sample and distribution quality in diffusion models, 2024.
\newblock URL \url{https://arxiv.org/abs/2404.07724}.

\bibitem[Liu et~al.(2022)Liu, Li, Du, Torralba, and Tenenbaum]{liu2022compositional}
Nan Liu, Shuang Li, Yilun Du, Antonio Torralba, and Joshua~B Tenenbaum.
\newblock Compositional visual generation with composable diffusion models.
\newblock In \emph{European Conference on Computer Vision}, pages 423--439. Springer, 2022.

\bibitem[Morozov et~al.(2021)Morozov, Voynov, and Babenko]{morozov2021self}
Stanislav Morozov, Andrey Voynov, and Artem Babenko.
\newblock On self-supervised image representations for gan evaluation.
\newblock In \emph{International Conference on Learning Representations}, 2021.

\bibitem[Nam et~al.(2024)Nam, Kim, Lee, Jin, Kim, and Chang]{nam2024dreammatcher_att}
Jisu Nam, Heesu Kim, DongJae Lee, Siyoon Jin, Seungryong Kim, and Seunggyu Chang.
\newblock Dreammatcher: Appearance matching self-attention for semantically-consistent text-to-image personalization.
\newblock In \emph{Proceedings of the IEEE/CVF Conference on Computer Vision and Pattern Recognition}, pages 8100--8110, 2024.

\bibitem[Oquab et~al.(2023)Oquab, Darcet, Moutakanni, Vo, Szafraniec, Khalidov, Fernandez, Haziza, Massa, El-Nouby, Howes, Huang, Xu, Sharma, Li, Galuba, Rabbat, Assran, Ballas, Synnaeve, Misra, Jegou, Mairal, Labatut, Joulin, and Bojanowski]{oquab2023dinov2}
Maxime Oquab, Timothée Darcet, Theo Moutakanni, Huy~V. Vo, Marc Szafraniec, Vasil Khalidov, Pierre Fernandez, Daniel Haziza, Francisco Massa, Alaaeldin El-Nouby, Russell Howes, Po-Yao Huang, Hu~Xu, Vasu Sharma, Shang-Wen Li, Wojciech Galuba, Mike Rabbat, Mido Assran, Nicolas Ballas, Gabriel Synnaeve, Ishan Misra, Herve Jegou, Julien Mairal, Patrick Labatut, Armand Joulin, and Piotr Bojanowski.
\newblock Dinov2: Learning robust visual features without supervision, 2023.

\bibitem[Owen(1980)]{owen1980integrals}
D.~B. Owen.
\newblock A table of normal integrals.
\newblock \emph{Communications in Statistics - Simulation and Computation}, 9\penalty0 (4):\penalty0 389--419, 1980.
\newblock \doi{10.1080/03610918008812164}.
\newblock URL \url{https://doi.org/10.1080/03610918008812164}.

\bibitem[Parmar et~al.(2022)Parmar, Zhang, and Zhu]{parmar2022aliasedresizingsurprisingsubtleties}
Gaurav Parmar, Richard Zhang, and Jun-Yan Zhu.
\newblock On aliased resizing and surprising subtleties in gan evaluation, 2022.
\newblock URL \url{https://arxiv.org/abs/2104.11222}.

\bibitem[Peebles and Xie(2023)]{peebles2023scalable_dit}
William Peebles and Saining Xie.
\newblock Scalable diffusion models with transformers.
\newblock In \emph{Proceedings of the IEEE/CVF International Conference on Computer Vision}, pages 4195--4205, 2023.

\bibitem[Peli(1990)]{peli1990contrast}
Eli Peli.
\newblock Contrast in complex images.
\newblock \emph{Journal of the Optical Society of America A}, 7\penalty0 (10):\penalty0 2032--2040, 1990.

\bibitem[Rombach et~al.(2022)Rombach, Blattmann, Lorenz, Esser, and Ommer]{rombach2022high}
Robin Rombach, Andreas Blattmann, Dominik Lorenz, Patrick Esser, and Bj{\"o}rn Ommer.
\newblock High-resolution image synthesis with latent diffusion models.
\newblock In \emph{Proceedings of the IEEE/CVF conference on computer vision and pattern recognition}, pages 10684--10695, 2022.

\bibitem[Ruhe et~al.(2024)Ruhe, Heek, Salimans, and Hoogeboom]{ruhe2024rolling_sliding}
David Ruhe, Jonathan Heek, Tim Salimans, and Emiel Hoogeboom.
\newblock Rolling diffusion models.
\newblock \emph{arXiv preprint arXiv:2402.09470}, 2024.

\bibitem[Sadat et~al.()Sadat, Hilliges, and Weber]{sadat2024eliminating}
Seyedmorteza Sadat, Otmar Hilliges, and Romann~M Weber.
\newblock Eliminating oversaturation and artifacts of high guidance scales in diffusion models.
\newblock In \emph{The Thirteenth International Conference on Learning Representations}.

\bibitem[Sadat et~al.(2024{\natexlab{a}})Sadat, Buhmann, Bradley, Hilliges, and Weber]{sadat2024cads}
Seyedmorteza Sadat, Jakob Buhmann, Derek Bradley, Otmar Hilliges, and Romann~M. Weber.
\newblock Cads: Unleashing the diversity of diffusion models through condition-annealed sampling.
\newblock In \emph{ICLR}, 2024{\natexlab{a}}.
\newblock URL \url{https://openreview.net/forum?id=zMoNrajk2X}.

\bibitem[Sadat et~al.(2024{\natexlab{b}})Sadat, Kansy, Hilliges, and Weber]{sadat2024icg}
Seyedmorteza Sadat, Manuel Kansy, Otmar Hilliges, and Romann~M Weber.
\newblock No training, no problem: Rethinking classifier-free guidance for diffusion models.
\newblock \emph{arXiv preprint arXiv:2407.02687}, 2024{\natexlab{b}}.

\bibitem[Saharia et~al.(2022)Saharia, Chan, Saxena, Li, Whang, Denton, Ghasemipour, Gontijo~Lopes, Karagol~Ayan, Salimans, et~al.]{saharia2022photorealistic}
Chitwan Saharia, William Chan, Saurabh Saxena, Lala Li, Jay Whang, Emily~L Denton, Kamyar Ghasemipour, Raphael Gontijo~Lopes, Burcu Karagol~Ayan, Tim Salimans, et~al.
\newblock Photorealistic text-to-image diffusion models with deep language understanding.
\newblock \emph{Advances in neural information processing systems}, 35:\penalty0 36479--36494, 2022.

\bibitem[Salimans and Ho(2022)]{salimans2022progressivedistillationfastsampling}
Tim Salimans and Jonathan Ho.
\newblock Progressive distillation for fast sampling of diffusion models, 2022.
\newblock URL \url{https://arxiv.org/abs/2202.00512}.

\bibitem[Shao et~al.(2024)Shao, Yang, Zhou, Zhang, Shen, Poggi, and Liao]{shao2024learning}
Jiahao Shao, Yuanbo Yang, Hongyu Zhou, Youmin Zhang, Yujun Shen, Matteo Poggi, and Yiyi Liao.
\newblock Learning temporally consistent video depth from video diffusion priors.
\newblock \emph{arXiv preprint arXiv:2406.01493}, 2024.

\bibitem[Sohl-Dickstein et~al.(2015)Sohl-Dickstein, Weiss, Maheswaranathan, and Ganguli]{sohldickstein2015deep}
Jascha Sohl-Dickstein, Eric~A. Weiss, Niru Maheswaranathan, and Surya Ganguli.
\newblock Deep unsupervised learning using nonequilibrium thermodynamics, 2015.

\bibitem[Song et~al.(2021)Song, Sohl-Dickstein, Kingma, Kumar, Ermon, and Poole]{song2021scorebased}
Yang Song, Jascha Sohl-Dickstein, Diederik~P. Kingma, Abhishek Kumar, Stefano Ermon, and Ben Poole.
\newblock Score-based generative modeling through stochastic differential equations, 2021.

\bibitem[Stein et~al.(2023)Stein, Cresswell, Hosseinzadeh, Sui, Ross, Villecroze, Liu, Caterini, Taylor, and Loaiza-Ganem]{stein2023exposingflawsgenerativemodel}
George Stein, Jesse~C. Cresswell, Rasa Hosseinzadeh, Yi~Sui, Brendan~Leigh Ross, Valentin Villecroze, Zhaoyan Liu, Anthony~L. Caterini, J.~Eric~T. Taylor, and Gabriel Loaiza-Ganem.
\newblock Exposing flaws of generative model evaluation metrics and their unfair treatment of diffusion models, 2023.
\newblock URL \url{https://arxiv.org/abs/2306.04675}.

\bibitem[Tang et~al.(2025)Tang, Bao, Chen, and Guo]{tang2025diffusion}
Zhicong Tang, Jianmin Bao, Dong Chen, and Baining Guo.
\newblock Diffusion models without classifier-free guidance.
\newblock \emph{arXiv preprint arXiv:2502.12154}, 2025.

\bibitem[Von~Luxburg and Sch{\"o}lkopf(2011)]{von2011statistical}
Ulrike Von~Luxburg and Bernhard Sch{\"o}lkopf.
\newblock Statistical learning theory: Models, concepts, and results.
\newblock In \emph{Handbook of the History of Logic}, volume~10, pages 651--706. Elsevier, 2011.

\bibitem[Wang et~al.(2024)Wang, Dufour, Andreou, Cani, Abrevaya, Picard, and Kalogeiton]{wang2024analysis_cfg}
Xi~Wang, Nicolas Dufour, Nefeli Andreou, Marie-Paule Cani, Victoria~Fern{\'a}ndez Abrevaya, David Picard, and Vicky Kalogeiton.
\newblock Analysis of classifier-free guidance weight schedulers.
\newblock \emph{arXiv preprint arXiv:2404.13040}, 2024.

\end{thebibliography}
